\documentclass[10pt,journal,compsoc]{IEEEtran}
\ifCLASSOPTIONcompsoc
  \usepackage[nocompress]{cite}
\else
  \usepackage{cite}
\fi
\ifCLASSINFOpdf
  \usepackage[pdftex]{graphicx}
  \graphicspath{{../pdf/}{../jpeg/}}
  \DeclareGraphicsExtensions{.pdf,.jpeg,.png}
\else
  \usepackage[dvips]{graphicx}
  \graphicspath{{../eps/}}
  \DeclareGraphicsExtensions{.eps}
\fi
\usepackage{amsmath}
\usepackage{array}
\usepackage{url}

\usepackage{graphicx}
\usepackage{subfigure}

\usepackage{amssymb}

\usepackage[breaklinks=true,colorlinks,bookmarks=false]{hyperref}

\usepackage{algorithm}
\usepackage[noend]{algorithmic}

\usepackage{bm}

\usepackage{multicol}
\usepackage{multirow}

\usepackage{threeparttable}

\def\methodname{PCL}

\DeclareMathOperator*{\argmax}{argmax}

\newcolumntype{L}[1]{>{\raggedright\let\newline\\\arraybackslash\hspace{0pt}}m{#1}}
\newcolumntype{R}[1]{>{\raggedleft\let\newline\\\arraybackslash\hspace{0pt}}m{#1}}
\newcolumntype{C}[1]{>{\centering\let\newline\\\arraybackslash\hspace{0pt}}m{#1}}

\newcolumntype{x}{>\small c}

\def\eg{\emph{e.g}.} 
\def\ie{\emph{i.e}.} 
 
\def\etc{\emph{etc}.} 
\def\wrt{w.r.t.} 
\def\etal{\emph{et al}.}

\usepackage{color}

\begin{document}
%
\title{PCL: Proposal Cluster Learning for Weakly Supervised Object Detection}
%
%
%
%

\author{Peng~Tang,
        Xinggang~Wang,~\IEEEmembership{Member,~IEEE,}
        Song~Bai,
        Wei~Shen,
        Xiang~Bai,~\IEEEmembership{Senior~Member,~IEEE,}
        Wenyu~Liu,~\IEEEmembership{Senior~Member,~IEEE,}
        and~Alan~Yuille,~\IEEEmembership{Fellow,~IEEE}
\IEEEcompsocitemizethanks{
\IEEEcompsocthanksitem P. Tang, X. Wang, X. Bai, and W. Liu are with the School of Electronic Information and Communications, Huazhong University of Science and Technology, Wuhan, 430074 China.\protect\\
E-mail: \{pengtang, xgwang, xbai, liuwy\}@hust.edu.cn
}
\IEEEcompsocitemizethanks{
\IEEEcompsocthanksitem S. Bai is with the Department of Engineering Science, University of Oxford, Oxford, OX1 3PJ, UK.\protect\\
E-mail: songbai.site@gmail.com
}
\IEEEcompsocitemizethanks{
\IEEEcompsocthanksitem W. Shen is with the Key Laboratory of Specialty Fiber Optics and Optical Access Networks, Shanghai Institute for Advanced Communication and Data Science, School of Communication and Information Engineering, Shanghai University, Shanghai, 200444 China.\protect\\
E-mail: shenwei1231@gmail.com
}
\IEEEcompsocitemizethanks{
\IEEEcompsocthanksitem A. Yuille is with the Departments of Cognitive Science and Computer Science, Johns Hopkins University, Baltimore, MD 21218-2608 USA.\protect\\
E-mail: alan.l.yuille@gmail.com
}
}

%
%

\markboth{Journal of \LaTeX\ Class Files}%
{Tang \MakeLowercase{\textit{et al.}}: PCL: Proposal Cluster Learning for Weakly Supervised Object Detection}
%



\IEEEtitleabstractindextext{%

\begin{abstract}
    Weakly Supervised Object Detection (WSOD),
    using only image-level annotations to train object detectors,
    is of growing importance in object recognition.
    In this paper, we propose a novel deep network for WSOD.
    Unlike previous networks that transfer the object detection problem to an image classification problem
    using Multiple Instance Learning (MIL),
    our strategy generates proposal clusters to learn refined instance classifiers by an iterative process.
    The proposals in the same cluster are spatially adjacent and associated with the same object.
    This prevents the network from concentrating too much on parts of objects instead of whole objects.
    We first show that instances can be assigned object or background labels directly based on proposal clusters for instance classifier refinement,
    and then show that treating each cluster as a small new bag yields fewer ambiguities than the directly assigning label method.
    The iterative instance classifier refinement is implemented online using multiple streams in convolutional neural networks,
    where the first is an MIL network and the others are for instance classifier refinement supervised by the preceding one.
    Experiments are conducted on the PASCAL VOC, ImageNet detection, and {MS-COCO} benchmarks for WSOD.
    Results show that our method outperforms the previous state of the art significantly.
\end{abstract}

}

\maketitle

\IEEEdisplaynontitleabstractindextext

%
\IEEEpeerreviewmaketitle

\IEEEraisesectionheading{
\section{Introduction}
\label{sec:intro}}
\IEEEPARstart{O}{bject} detection is one of the most important problems in computer vision with many applications.
Recently, due to the development of Convolutional Neural Network (CNN) \cite{Ref:Lecun1998,Ref:Krizhevsky2012} and the availability of large scale datasets with detailed boundingbox-level annotations \cite{Ref:Everingham2015,Ref:Russakovsky2015,Ref:Lin2014},
there have been great leap forwards in object detection \cite{Ref:Girshick2016,Ref:Girshick2015,Ref:Ren2017,Ref:Redmon2016,Ref:Liu2016,Ref:Zhang2018}.
However, it is very labor-intensive and time-consuming to collect detailed annotations,
whereas acquiring images with only image-level annotations (\ie, image tags) indicating whether an object class exists in an image or not is much easier.
For example, we can use image search queries to search on the Internet (\eg, Google and Flickr) to obtain a mass of images with such image-level annotations.
This fact inspires us to explore methods for the Weakly Supervised Object Detection (WSOD) problem, \ie, training object detectors with only image tag supervisions.

Many previous methods follow the Multiple Instance Learning (MIL) pipeline for WSOD \cite{Ref:Ren2016,Ref:Cinbis2017,Ref:Wang2015,Ref:Shi2017,Ref:Tang2017deep,Ref:Bilen2016,Ref:Kantorov2016,Ref:Diba2017}.
They treat images as bags and proposals as instances;
then instance classifiers (object detectors) are trained under MIL constraints
(\ie, a positive bag contains at least one positive instance and all instances in negative bags are negative).
In addition, inspired by the great success of CNN,
recent efforts often combine MIL and CNN to obtain better WSOD performance.
Some researches have shown that treating CNNs pre-trained on large scale datasets as off-the-shelf proposal feature extractors can obtain much better performance than traditional hand-designed features \cite{Ref:Ren2016,Ref:Cinbis2017,Ref:Wang2015,Ref:Shi2017}.
{Moreover, many recent works have achieved even better results for WSOD by an MIL network using standard end-to-end training \cite{Ref:Tang2017deep,Ref:Kantorov2016} or a variant of end-to-end training \cite{Ref:Bilen2016,Ref:Diba2017}.
See Section~\ref{sec:eav} for this variant of end-to-end and how it differs from the standard one.
We use the same strategy of training a variant of end-to-end MIL network inspired by \cite{Ref:Bilen2016,Ref:Diba2017}.}

\begin{figure}[t]
\begin{center}
   \includegraphics[width=0.965\linewidth]{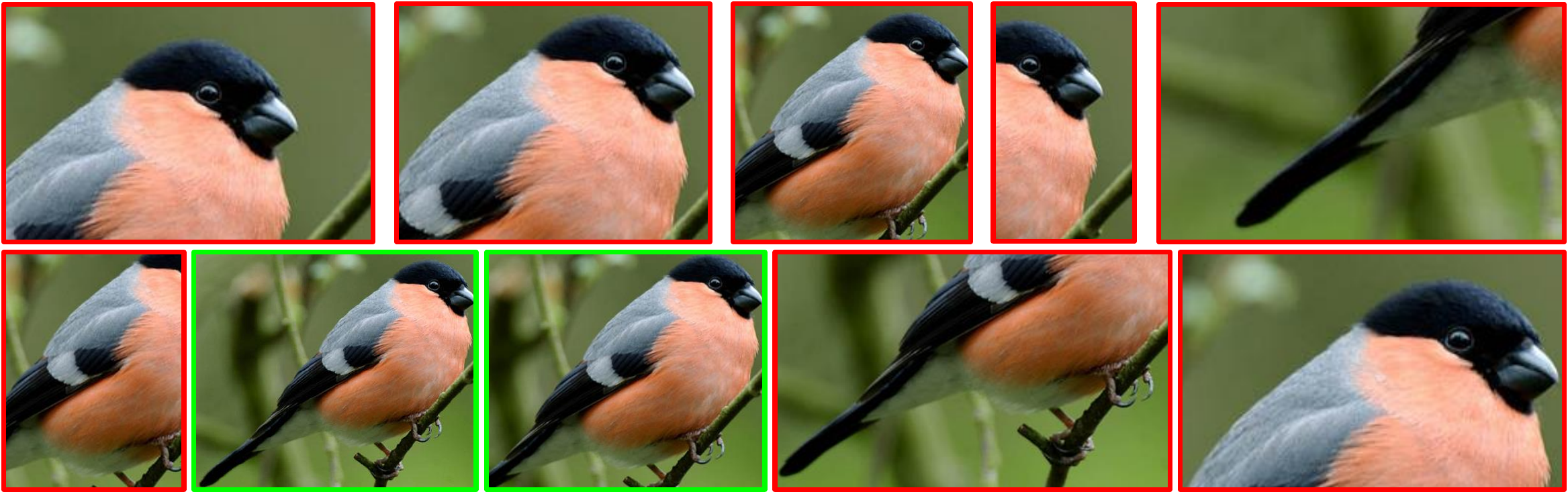}
\end{center}
   \caption{Different proposals cover different parts of objects.
   All these proposals can be classified as ``bird''
   but only the green boxes, which have enough IoU with groundtruth, contribute to correct detections.
   }
\label{fig:bird}
\end{figure}

\begin{figure}[t]
\begin{center}
   \includegraphics[width=0.98\linewidth]{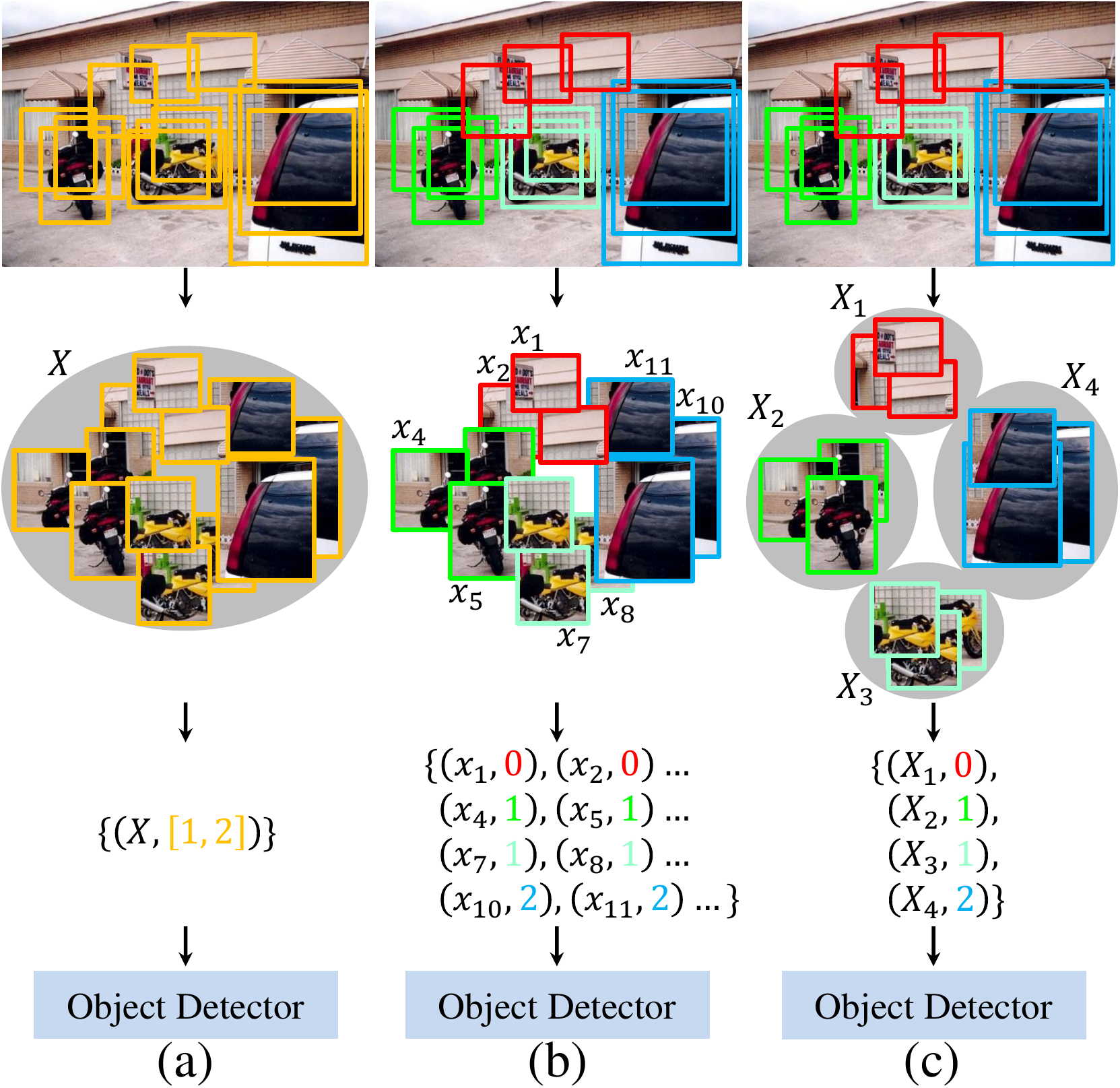}
\end{center}
   \caption{(a) Conventional MIL networks transfer the instance classification (object detection) problem to a bag classification (image classification) problem.
   (b) We propose to generate proposal clusters and assign proposals the label of the corresponding object class for each cluster.
   (c) We propose to treat each proposal cluster as a small new bag.
   ``0'', ``1'', and ``2'' indicate the ``background'', ``motorbike'', and ``car'', respectively.
   }
\label{fig:method_compare}
\end{figure}

Although some promising results have been obtained by MIL networks for WSOD,
they do not perform as well as fully supervised ones \cite{Ref:Girshick2016,Ref:Girshick2015,Ref:Ren2017}.
As shown in Fig.~\ref{fig:method_compare}~(a),
previous MIL networks integrate the MIL constraints into the network training by transferring the instance classification (object detection) problem to a bag classification (image classification) problem,
where the final image scores are the aggregation of the proposal scores.
However, there is a big gap between image classification and object detection.
For classification, even parts of objects can contribute to correct results (\eg, the red boxes in Fig.~\ref{fig:bird}),
because important parts include many characteristics of the objects.
Many proposals only cover parts of objects,
and ``seeing'' proposals only of parts may be enough to roughly localize the objects.
But this may not localize objects well enough considering the performance requirement of high Intersection-over-Union (IoU) between the resulting boxes and groundtruth boundingboxes:
the top ranking proposals may only localize parts of objects instead of whole objects.
Recall that for detection, the resulting boxes should not only give correct classification,
but also localize objects and have enough overlap with groundtruth boundingboxes (\eg, the green boxes in Fig.~\ref{fig:bird}).

\begin{figure}[t]
\begin{center}
   \includegraphics[width=0.965\linewidth]{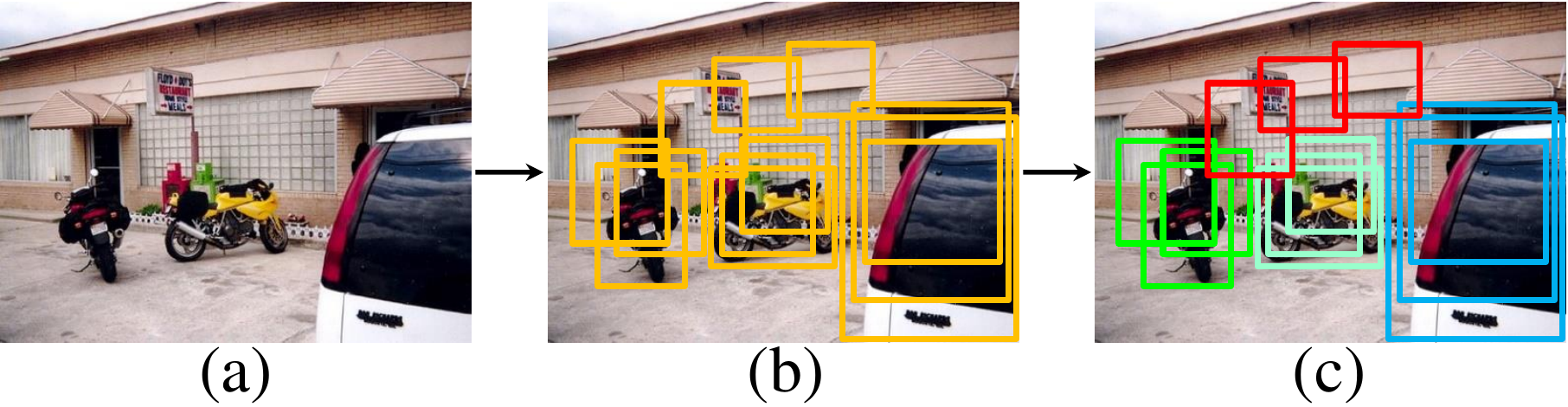}
\end{center}
   \caption{The proposals (b), of an image (a), can be grouped into different proposal clusters (c).
   Proposals with the same color in (c) belong to the same cluster (red indicates background).}
\label{fig:proposal_cluster}
\end{figure}

Before presenting our solution of the problem referred above,
we first introduce the concept of proposal cluster.
Object detection requires algorithms to generate multiple overlapping proposals closely surrounding objects to ensure high proposal recall
(\eg, for each object, there are tens of proposals on average from Selective Search \cite{Ref:Uijlings2013} which have IoU$>$0.5 with the groundtruth boundingbox on the PASCAL VOC dataset).
Object proposals in an image can be grouped into different spatial clusters.
Except for one cluster for background proposals,
each object cluster is associated with a single object
and proposals in each cluster are spatially adjacent,
as shown in Fig.~\ref{fig:proposal_cluster}.
For fully supervised object detection
(\ie, training object detectors using boundingbox-level annotations), 
proposal clusters can be generated
by treating the groundtruth boundingboxes as cluster centers.
Then object detectors are trained according to the proposal clusters
(\eg, assigning all proposals the label of the corresponding object class for each cluster).
This alleviates the problem that detectors may only focus on parts.

But in the weakly supervised scenario,
it is difficult to generate proposal clusters
because groundtruth boundingboxes that can be used as cluster centers are not provided.
To cope with this difficulty, we suggest to find proposal clusters as follows.
First we generate proposal cluster centers
from those proposals which have high classification scores during training,
because these top ranking proposals can always detect at least parts of objects.
That is, for each image, after obtaining proposal scores, we select some proposals with high scores as cluster centers,
and then proposal clusters are generated based on spatial overlaps with the cluster centers.
Then the problem reduces to how to select proposals as centers,
because many high scoring proposals may correspond to the same object.
The most straightforward way is to choose the proposal with the highest score for each positive object class (\ie, the object class exists in the image) as the center.
But such a method ignores the fact that there may exist more than one object with the same object category in natural images (\eg, the two motorbikes in Fig.~\ref{fig:proposal_cluster}).
Therefore, we propose a graph-based method to find cluster centers.
More specifically, we build a graph of top ranking proposals according to the spatial similarity for each positive object class.
In the graph, two proposals are connected if they have enough spatial overlaps.
Then we greedily and iteratively choose the proposals which have most connections with others to estimate the centers.
Although a cluster center proposal may only capture an object partially,
its adjacent proposals (\ie, other proposals in the cluster) can cover the whole object,
or at worst contain larger parts of the object.

\begin{figure*}[!t]
\begin{center}
   \includegraphics[width=\linewidth]{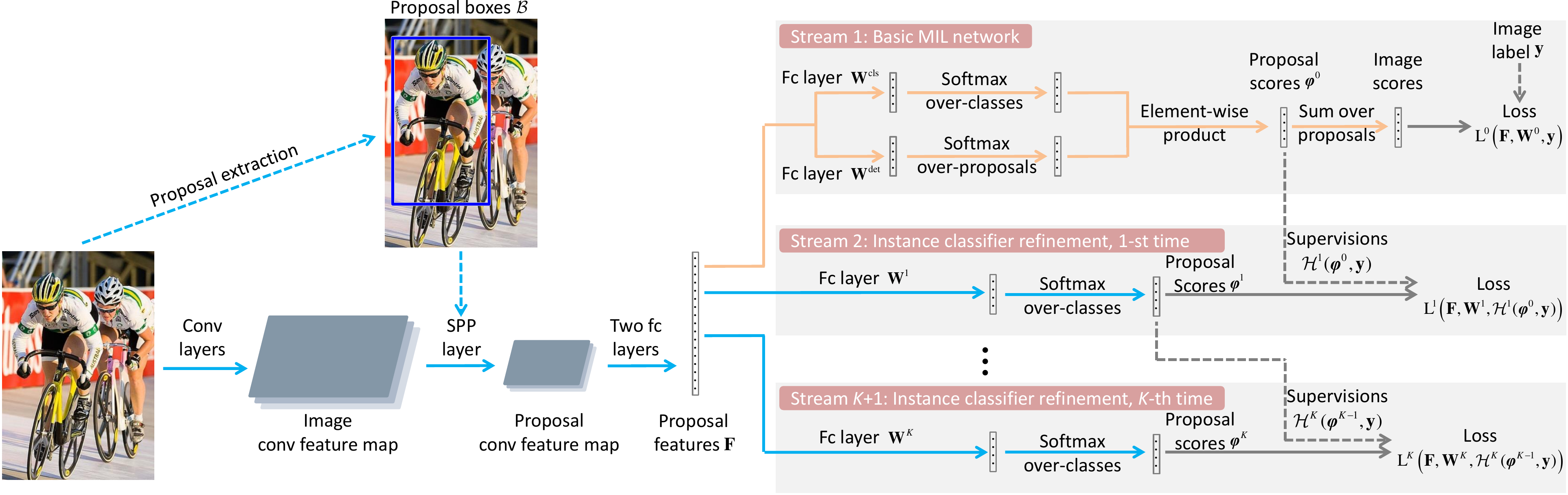}
\end{center}
   \caption{The architecture of our method.
   {All arrows are utilized during the forward process of training,
   only the solid ones have back-propagation computations,
   and only the blue ones are used during testing.
   During the forward process of training,
   an image and its proposal boxes are fed into the CNN
   which involves a series of convolutional layers, an SPP layer, and two fully connected layers
   to produce proposal features.
   These proposal features are branched into many streams:
   the first one for the basic MIL network
   and the other ones for iterative instance classifier refinement.
   Each stream outputs a set of proposal scores and generates proposal clusters consequently.
   Based on these proposal clusters,
   supervisions are generated to compute losses for the next stream.
   During the back-propagation process of training,
   proposal features and classifiers are trained according to the network losses.
   All streams share the same proposal features.}
   }
\label{fig:architecture}
\end{figure*}

Based on these proposal clusters, we propose two methods to refine instance classifiers (object detectors) during training.
We first propose to assign proposals object labels directly.
That is, for each cluster, we assign its proposals the label of its corresponding object class,
as in Fig.~\ref{fig:method_compare}~(b).
Compared with the conventional MIL network in Fig.~\ref{fig:method_compare}~(a),
this strategy forces network to ``see'' larger parts of objects by assigning object labels to proposals that cover larger parts of objects directly,
which fills the gap between classification and detection to some extent.
While effective, this strategy still has potential ambiguities,
because assigning the same object label to proposals that cover different parts of objects simultaneously may confuse the network
and will hurt the discriminative power of the detector.
To address this problem, we propose to treat each proposal cluster as a small new bag to train refined instance classifiers, as in Fig.~\ref{fig:method_compare}~(c).
Most of the proposals in these new bags should have relatively high classification scores
because the cluster centers covers at least parts of objects and proposals in the same cluster are spatially adjacent (except for the background cluster).
In the same time,
not all proposals in the bags should have high classification scores.
Thus compared with the directly assigning label strategy, this strategy is more flexible and can reduce the ambiguities to some extent.
We name our method Proposal Cluster Learning (\methodname) because it learns refined instance classifiers based on proposal clusters.

To implement our idea effectively and efficiently, we further propose an online training approach.
Our network has multiple output streams as in Fig.~\ref{fig:architecture}.
The first stream is a basic MIL network which aggregates proposal scores into final image scores to train basic instance classifiers,
and the other streams refine the instance classifiers iteratively.
{During the forward process of training, proposal classification scores are obtained and proposal clusters are generated consequently for each stream.
Then based on these proposal clusters, supervisions are generated to compute losses for the next stream.
According to the losses, these refined classifiers are trained during back-propagation.}
Except for the first stream that is supervised by image labels,
the other streams are supervised by the image labels as well as outputs from their preceding streams.
As our method forces the network to ``see'' larger parts of objects,
the detector can discover the whole object instead of parts gradually by performing refinement multiple times (\ie, multiple output streams).
But at the start of training, all classifiers are almost untrained,
which will result in very noisy proposal clusters,
and so the training will deviate from the correct solutions a lot.
Thus we design a weighted loss further by associating different proposals with different weights in different training iterations.
After that, all training procedures can thus be integrated into a single end-to-end network.
This can improve the performance benefiting from our \methodname-based classifier refinement procedure.
It is also very computational efficient in both training and testing.
In addition, performance can be improved by sharing proposal features among different output streams.

We elaborately conduct many experiments on the challenging PASCAL VOC, ImageNet detection, {and MS-COCO} datasets to confirm the effectiveness of our method.
Our method achieves $48.8\%$ mAP and $66.6\%$ CorLoc on VOC 2007 which is more than $5\%$ absolute improvement compared with previous best performed methods.

This paper is an extended version of our previous work \cite{Ref:Tang2017multiple}.
In particular, we give more analyses of our method
and enrich literatures of most recent related works,
making the manuscript more complete.
In addition, we make two methodological improvements:
the first one is to generate proposal clusters using graphs of top ranking proposals instead of using the highest scoring proposal,
and the second one is to treat each proposal cluster as a small new bag.
In addition, we provide more discussions of experimental results,
and show the effectiveness of our method on the challenging ImageNet detection {and MS-COCO} datasets.

The rest of our paper is organized as follows.
In Section~\ref{sec:related_work}, some related works are introduced.
In Section~\ref{sec:method}, the details of our method are described.
Elaborate experiments and analyses are conducted in Section~\ref{sec:exp}.
Finally, conclusions and future directions are presented in Section~\ref{sec:conclu}.

\section{Related work}
\label{sec:related_work}

\subsection{Multiple instance learning}
\label{sec:rea_mil}

MIL, first proposed for drug activity prediction \cite{Ref:Dietterich1997}, is a classical weakly supervised learning problem.
Many variants have been proposed for MIL \cite{Ref:Andrews2003,Ref:Wang2015,Ref:Zhang2002,Ref:Wang2018}.
In MIL, a set of bags are given, and each bag is associated with a collection of instances.
It is natural to treat WSOD as an MIL problem.
Then the problem turns into finding instance classifiers only given bag labels.
Our method also follows the MIL strategy and makes several improvements to WSOD.
In particular, we learn refined instance classifiers based on proposal clusters according to both instance scores and spatial relations in an online manner.
\footnote{``Instance'' and ``proposal'' are used interchangeably in this paper.}

MIL has many applications to computer vision, such as image classification \cite{Ref:Li2013,Ref:Tang2017learning}, weakly supervised semantic segmentation \cite{Ref:Pathak15,Ref:Pinheiro2015}, object detection \cite{Ref:Zhang2006}, object tracking \cite{Ref:Babenko2011}, \etc\
The strategy of treating proposal clusters as bags was partly inspired by \cite{Ref:Zhang2006,Ref:Babenko2011},
where \cite{Ref:Zhang2006} proposes to train MIL for patches around groundtruth locations
and \cite{Ref:Babenko2011} proposes to train MIL for patches around predicted object locations.
However, they require groundtruth locations for either all training samples \cite{Ref:Zhang2006} or the beginning time frames \cite{Ref:Babenko2011},
whereas WSOD does not have such annotations.
Therefore, it is much harder to generate proposal clusters only guided by image-level supervisions for WSOD.
In addition, we incorporate the strategy of treating proposal clusters as bags into the network training whereas \cite{Ref:Zhang2006,Ref:Babenko2011} do not.
Oquab \etal\ \cite{Ref:Oquab2015} also train a CNN network using the max pooing MIL strategy to localize objects.
But their methods can only coarsely localize objects regardless of their sizes and aspect ratios, whereas our method can detect objects more accurately.

\subsection{Weakly supervised object detection}
\label{sec:wsod}

WSOD has attracted great interests nowadays because the amount of data with image-level annotations is much bigger and is growing much faster than that with boundingbox-level annotations.
Many methods are emerging for the WSOD problem \cite{Ref:Chum2007,Ref:Deselaers2012,Ref:Pandey2011,Ref:Shi2015,Ref:Song2014learning,Ref:Song2014,Ref:Cinbis2017,Ref:Bilen2015,Ref:Wang2015,Ref:Wang2014}.
For example, Chum and Zisserman \cite{Ref:Chum2007} first initialize object locations by discriminative visual words and then introduce an exemplar model to measure similarity between image pairs for updating locations.
Deselaers \etal\ \cite{Ref:Deselaers2012} propose to initialize boxes by objectness \cite{Ref:Alexe2012} and use a CRF-based model to iteratively localize objects.
Pandey and Lazebnik \cite{Ref:Pandey2011} train a DPM model \cite{Ref:Felzenszwalb2010} under weak supervisions for WSOD.
Shi \etal\ \cite{Ref:Shi2015} use Bayesian latent topic models to jointly model different object classes and background.
Song \etal\ \cite{Ref:Song2014} develop a technology to discover frequent discriminative configurations of visual patterns for robust WSOD.
Cinbis \etal\ \cite{Ref:Cinbis2017} iteratively train a multi-fold MIL to avoid the detector being locked onto inaccurate local optima.
Wang \etal\ \cite{Ref:Wang2015} relax the MIL constraints into a derivable loss function to train detectors more efficient.

Recently, with the revolution of CNNs in computer vision, many works also try to combine the WSOD with CNNs.
Early works treat CNN models pre-trained on ImageNet as off-the-shelf feature extractors \cite{Ref:Song2014learning,Ref:Song2014,Ref:Cinbis2017,Ref:Bilen2015,Ref:Wang2015,Ref:Wang2014,Ref:Ren2016,Ref:Shi2017}.
They extract CNN features for each candidate regions, and then train their own detectors on top of these features.
These methods have shown that CNN descriptors can boost performance against traditional hand-designed features.
More recent efforts tend to train end-to-end networks for WSOD \cite{Ref:Tang2017deep,Ref:Bilen2016,Ref:Kantorov2016,Ref:Diba2017}.
They integrate the MIL constraints into the network training by aggregating proposal classification scores into final image classification scores,
and then image-level supervision can be directly added to image classification scores.
For example, Tang \etal\ \cite{Ref:Tang2017deep} propose to use max pooling for aggregation.
Bilen and Vedaldi \cite{Ref:Bilen2016} develop a weighted sum pooing strategy.
Building on \cite{Ref:Bilen2016}, Kantorov \etal\ argue that context information can improve the performance.
Diba \etal\ \cite{Ref:Diba2017} show that weakly supervised segmentation map can be used as guidance to filter proposals,
and jointly train the weakly supervised segmentation network and WSOD end-to-end.
Our method is built on these networks and any of them can be chosen as our basic network.
Our strategy proposes to learn refined instance classifiers based on proposal clusters,
and propose a novel online approach to train our network effectively and efficiently.
Experimental results show our strategies can boost the results significantly.

In addition to the weighted sum pooing,
\cite{Ref:Bilen2016} also proposes a ``spatial regulariser'' that forces features of the highest scoring proposal and its spatially adjacent proposals to be the same.
Unlike this, we show that finding proposal cluster centers using graph
and treating proposal clusters as bags are more effective.
The contemporary work \cite{Ref:Jie2017} uses a graph model to generate seed proposals.
Their network training has many steps:
first, an MIL network \cite{Ref:Wei2016} is trained;
second, seed proposals are generated using the graph;
third, based on these seed proposals, a Fast R-CNN \cite{Ref:Girshick2015} like detector is trained.
Our method differs from \cite{Ref:Jie2017} in many aspects:
first, we propose to generate proposal clusters for each training iteration
and thus our network is trained end-to-end instead of step-by-step,
which is more efficient and can benefit from sharing proposal features among different streams;
second, we propose to treat proposal clusters as bags for training better classifiers.
As evidenced by experiments, our method obtains much better and more robust results.

{
\subsection{End-to-end and its variants}
\label{sec:eav}
In standard end-to-end training,
the update requires optimizing losses \wrt\ all functions of network parameters.
For example,
the Fast R-CNN \cite{Ref:Girshick2015} optimizes their classification loss and boundingbox regression loss \wrt\ proposal classification and feature extraction for fully supervised object detection.
The MIL networks in \cite{Ref:Tang2017deep,Ref:Kantorov2016} optimize their MIL loss \wrt\ proposal classification and feature extraction for WSOD.

Unlike the standard end-to-end training,
there exists a variant of end-to-end training.
The variant contains functions
which depend on network parameters,
but losses are not optimized \wrt\ all these functions \cite{Ref:Bilen2016,Ref:Diba2017}.
As we described in Section~\ref{sec:wsod},
the ``spatial regulariser'' in \cite{Ref:Bilen2016}
forces features of the highest scoring proposal and its spatially adjacent proposals to be the same.
They use a function of network parameters to compute the highest scoring proposal,
and do not optimize their losses \wrt\ this function.
Diba \etal\ \cite{Ref:Diba2017} filter out background proposals using a function of network parameters
and use these filtered proposals in their latter network computations.
They also do not optimize their losses \wrt\ this function.
Inspired by \cite{Ref:Bilen2016,Ref:Diba2017},
we use this variant of end-to-end training.
More precisely,
we do not optimize our losses \wrt\ the generated supervisions for instance classifier refinement.
}

\subsection{Others}

There are many other important related works that do not focus on weakly supervised learning but should be discussed.
Similar to other end-to-end MIL networks,
our method is built on top of the Region of Interest (RoI) pooling layer \cite{Ref:Girshick2015} or Spatial Pyramid Pooling (SPP) layer \cite{Ref:He2015}
to share convolutional computations among different proposals for model acceleration.
But both \cite{Ref:Girshick2015} and \cite{Ref:He2015} require boundingbox-level annotations to train their detectors.
The sharing proposal feature strategy in our network is similar to multi-task learning \cite{Ref:Caruana1997}.
Unlike the multi-task learning that each output stream has their own relatively independent external supervisions for different tasks,
in our method, all streams have the same task and supervisions of later streams depend on the outputs from their preceding streams.

\section{Method}
\label{sec:method}

The overall architecture of our method is shown in Fig.~\ref{fig:architecture}.
Given an image, about $2,000$ object proposals from Selective Search \cite{Ref:Uijlings2013} or EdgeBox \cite{Ref:Zitnick2014} are generated.
{During the forward process of training,
the image and these proposals are fed into some convolutional (conv) layers with an SPP layer \cite{Ref:He2015} to produce a fixed-size conv feature map per-proposal.
After that, proposal feature maps are fed into two fully connected (fc) layers to produce proposal features.
These features are branched into different streams:
the first one is an MIL network to train basic instance classifiers and the others refine the classifiers iteratively.
For each stream, proposal classification scores are obtained and proposal clusters are generated consequently.
Then based on these proposal clusters,
supervisions are generated to compute losses for the next stream.
During the back-propagation process of training,
the network losses are optimized to train proposal features and classifiers.}
As shown in the figure, supervisions of the $1$-st refined classifier depend on the output from the basic classifier,
and supervisions of $k$-th refined classifier depend on outputs from $\{k-1\}$-th refined classifier.
In this section, we will introduce our method of learning refined instance classifiers based on proposal clusters in detail.

\subsection{Notations}
\label{sec:notation}

Before presenting our method,
we first introduce some of the mostly used notations as follows.
We have $R$ proposals with boxes ${\cal B} = \{b_{r}\}_{r=1}^{R}$ for an given image
and proposal features $\mathbf{F}$,
where $b_{r}$ is the $r$-th proposal box.
The number of refined instance classifiers is $K$ (\ie, we refine instance classifier $K$ times), and thus there are $K+1$ streams.
The number of object classes is $C$.
$\mathbf{W}^{0}$ and $\mathbf{W}^{k}, k \in \{1, ..., K\}$ are the parameters of the basic instance classifier and the $k$-th refined instance classifier, respectively.
$\bm{\varphi}^{0}(\mathbf{F}, \mathbf{W}^{0}) \in \mathbb{R}^{C \times R}$ and $\bm{\varphi}^{k}(\mathbf{F}, \mathbf{W}^{k}) \in \mathbb{R}^{(C+1) \times R}, k \in \{1, ..., K\}$ are the predicted score matrices
of the basic instance classifier and the $k$-th refined instance classifier, respectively,
where $C+1$ indicates the $C$ object classes and $1$ background class.
We use $\bm{\varphi}^{k}$ later for simplification, dropping the dependence on $\mathbf{F}, \mathbf{W}^{k}$.
$\varphi^{k}_{cr}$ is the predicted score of the $r$-th proposal for class $c$ from the $k$-th instance classifier.
$\mathbf{y} = [y_{1}, ..., y_{C}]^{T}$ is the image label vector,
where $y_{c}=1$ or $0$ indicates the image with or without object class $c$.
${\cal H}^{k}(\bm{\varphi}^{k-1}, \mathbf{y})$ is the supervision of the $k$-th instance classifier,
where ${\cal H}^{k}(\bm{\varphi}^{k-1}, \mathbf{y}), k=0$ is the image label vector $\mathbf{y}$.
$\textup{L}^{k}\left( \mathbf{F}, \mathbf{W}^{k}, {\cal H}^{k}(\bm{\varphi}^{k-1}, \mathbf{y}) \right)$ is the loss function to train the $k$-th instance classifier.

We compute $N^{k}$ proposal cluster centers ${\cal S}^{k} = \{S^{k}_{n}\}_{n=1}^{N^{k}}$ for the $k$-th refinement.
The $n$-th cluster center $S^{k}_{n} = (b^{k}_{n}, y^{k}_{n}, s^{k}_{n})$ consists of
a proposal box $b^{k}_{n} \in {\cal B}$,
an object label $y^{k}_{n}$ ($y^{k}_{n} = c, c \in \{1, ..., C\}$ indicates the $c$-th object class),
and a confidence score $s^{k}_{i}$ indicating the confidence that $b^{k}_{n}$ covers at least part of an object of class $y^{k}_{n}$.
We have $N^{k}+1$ proposal clusters ${\cal C}^{k} = \{{\cal C}^{k}_{n}\}_{n=1}^{N^{k}+1}$ according to ${\cal S}^{k}$
(${\cal C}^{k}_{N^{k}+1}$ for background and others for objects).
For object clusters,
the $n$-th cluster ${\cal C}^{k}_{n} = ({\cal B}^{k}_{n}, y^{k}_{n}, s^{k}_{n}), n \neq N^{k}+1$
consists of $M^{k}_{n}$ proposal boxes ${\cal B}^{k}_{n} = \{b^{k}_{nm}\}_{m=1}^{M^{k}_{n}} \subseteq {\cal B}$,
an object label $y^{k}_{n}$ that is the same as the cluster center label,
and a confidence score $s^{k}_{n}$ that is the same as the cluster center score,
where $s^{k}_{n}$ indicates the confidence that ${\cal C}^{k}_{n}$ corresponds to an object of class $y^{k}_{n}$.
Unlike object clusters,
the background cluster ${\cal C}^{k}_{n} = ({\cal P}^{k}_{n}, y^{k}_{n}), n = N^{k}+1$
consists of $M^{k}_{n}$ proposals ${\cal P}^{k}_{n} = \{P^{k}_{nm}\}_{m=1}^{M^{k}_{n}}$
and a label $y^{k}_{n} = C + 1$ indicating the background.
The $m$-th proposal $P^{k}_{nm} = (b^{k}_{nm}, s^{k}_{nm})$ consists of a proposal box $b^{k}_{nm} \in {\cal B}$
and a confidence score $s^{k}_{nm}$ indicating the confidence that $b^{k}_{nm}$ is the background.

\subsection{Basic MIL network}
\label{sec:mil}

It is necessary to generate proposal scores and clusters to supervise refined instance classifiers.
More specifically, the first refined classifier requires basic instance classifiers to generate proposal scores and clusters.
Therefore, we first introduce our basic MIL network as the basic instance classifier.
Our overall network is independent of the specific MIL methods,
and thus any method that can be trained end-to-end could be used.
There are many possible choices \cite{Ref:Tang2017deep,Ref:Bilen2016,Ref:Kantorov2016}.
Here we choose the method by Bilen and Vedaldi \cite{Ref:Bilen2016} which proposes a weighted sum pooling strategy to obtain the instance classifier,
because of its effectiveness and implementation convenience.
{To make our paper self-contained,
we briefly introduce \cite{Ref:Bilen2016} as follows.
}

Given an input image and its proposal boxes ${\cal B} = \{b_{r}\}_{r=1}^{R}$,
a set of proposal features $\mathbf{F}$ are first generated by the network.
Then as shown in the ``Basic MIL network'' block of Fig.~\ref{fig:architecture},
there are two branches which process the proposal features to produce two matrices $\mathbf{X}^{\textup{cls}}(\mathbf{F}, \mathbf{W}^{\textup{cls}}), \mathbf{X}^{\textup{det}}(\mathbf{F}, \mathbf{W}^{\textup{det}}) \in \mathbb{R}^{C \times R}$
(we use $\mathbf{X}^{\textup{cls}}, \mathbf{X}^{\textup{det}}$ later for simplification,
dropping the dependence on $\mathbf{F}, \mathbf{W}^{\textup{cls}}, \mathbf{W}^{\textup{det}}$)
of an input image by two fc layers,
where $\mathbf{W}^{\textup{cls}}$ and $\mathbf{W}^{\textup{det}}$
denote the parameters of the fc layer for $\mathbf{X}^{\textup{cls}}$
and the parameters of the fc layer for $\mathbf{X}^{\textup{det}}$, respectively.
Then the two matrices are passed through two softmax layer along different directions:
$[\bm{\sigma}(\mathbf{X}^{\textup{cls}})]_{cr} = {e^{x_{cr}^{\textup{cls}}}} / {\sum_{c^{\prime}=1}^{C} e^{x_{c^{\prime}r}^{\textup{cls}}}}$
and $[\bm{\sigma}(\mathbf{X}^{\textup{det}})]_{cr} = {e^{x_{cr}^{\textup{det}}}} / {\sum_{r^{\prime}=1}^{R} e^{x_{cr^{\prime}}^{\textup{det}}}}$.
Let us denote $(\mathbf{W}^{\textup{cls}}, \mathbf{W}^{\textup{det}})$ by $\mathbf{W}^{0}$.
The proposal scores are generated by element-wise product $\bm{\varphi}^{0}(\mathbf{F}, \mathbf{W}^{0}) = \bm{\sigma}(\mathbf{X}^{\textup{cls}}) \odot \bm{\sigma}(\mathbf{X}^{\textup{det}})$.
Finally, the image score of the $c$-th class $[\bm{\phi}(\mathbf{F}, \mathbf{W}^{0})]_{c}$ is obtained by the sum over all proposals: $[\bm{\phi}(\mathbf{F}, \mathbf{W}^{0})]_{c} = \sum_{r=1}^{R} [\bm{\varphi}^{0}(\mathbf{F}, \mathbf{W}^{0})]_{cr}$.

A simple interpretation of the two branches framework is as follows.
$[\bm{\sigma}(\mathbf{X}^{\textup{cls}})]_{cr}$ is the probability of the $r$-th proposal belonging to class $c$.
$[\bm{\sigma}(\mathbf{X}^{\textup{det}})]_{cr}$ is the normalized weight that indicates the contribution of the $r$-th proposal to image being classified to class $i$.
So $[\bm{\phi}(\mathbf{F}, \mathbf{W}^{0})]_{c}$ is obtained by weighted sum pooling and falls in the range of $(0, 1)$.
Given the image label vector $\mathbf{y} = [y_{1}, ..., y_{C}]^{T}$.
We train the basic instance classifier
by optimizing the multi-class cross entropy loss Eq.~(\ref{equ:loss_base}) \wrt\ $\mathbf{F}, \mathbf{W}^{0}$.
\begin{equation}
\label{equ:loss_base}
\begin{aligned}
   \mathrm{L}^{0}\left(\mathbf{F}, \mathbf{W}^{0}, \mathbf{y}\right) = -\mathop{\sum} \limits_{c=1}^{C} \{&(1 - y_{c}) \log (1 - [\bm{\phi}(\mathbf{F}, \mathbf{W}^{0})]_{c})\\
   + &y_{c} \log [\bm{\phi}(\mathbf{F}, \mathbf{W}^{0})]_{c}\}.
\end{aligned}
\end{equation}

\subsection{The overall training strategy}
\label{sec:ots}

To refine instance classifiers iteratively,
we add multiple output streams in our network where each stream corresponds to a refined classifier,
as shown in Fig.~\ref{fig:architecture}.
We integrate the basic MIL network and the classifier refinement into an end-to-end network to learn the refined classifier online.
Unlike the basic instance classifier,
for an input image the output score matrix $\bm{\varphi}^{k}(\mathbf{F}, \mathbf{W}^{k})$ of the $k$-th refined classifier is a $\{C+1\} \times R$ matrix and is obtained by passing the proposal features through a single fc layer (with parameters $\mathbf{W}^{k}$) as well as a softmax over-classes layer,
\ie, $\bm{\varphi}^{k}(\mathbf{F}, \mathbf{W}^{k}) \in \mathbb{R}^{(C+1) \times R}, k \in \{1, 2, ..., K\}$,
as in the ``Instance classifier refinement'' blocks of Fig.~\ref{fig:architecture}.
Notice that we use the same proposal features $\mathbf{F}$ for all classifiers.
We use $\bm{\varphi}^{k}$ later for simplification,
dropping the dependence on $\mathbf{F}, \mathbf{W}^{k}$.

As we stated before, supervisions to train the $k$-th instance classifier are generated based on proposal scores $\bm{\varphi}^{k-1}$ and image label $\mathbf{y}$.
Thus we denote the supervisions by ${\cal H}^{k}(\bm{\varphi}^{k-1}, \mathbf{y})$.
{Then we train our overall network by optimizing the loss Eq.~(\ref{equ:loss_all}) \wrt\ $\mathbf{F}, \mathbf{W}^{k}$.
We do not optimize the loss \wrt\ ${\cal H}^{k}(\bm{\varphi}^{k-1}, \mathbf{y})$,
which means that the supervisions ${\cal H}^{k}(\bm{\varphi}^{k-1}, \mathbf{y})$ are only computed in the forward process
and we do not compute their gradients to train our network.
}
\begin{equation}
\label{equ:loss_all}
\begin{aligned}
   \mathop{\sum} \limits_{k=0}^{K} \mathrm{L}^{k}\left(\mathbf{F}, \mathbf{W}^{k}, {\cal H}^{k}(\bm{\varphi}^{k-1}, \mathbf{y})\right).
\end{aligned}
\end{equation}
The loss $\mathrm{L}^{k}\left(\mathbf{F}, \mathbf{W}^{k}, {\cal H}^{k}(\bm{\varphi}^{k-1}, \mathbf{y})\right), k > 0$ for the $k$-th refined instance classifier is defined in later Eq.~(\ref{equ:loss_label})/(\ref{equ:loss_label_weighted})/(\ref{equ:loss_bag})
which are loss functions with supervisions provided by ${\cal H}^{k}(\bm{\varphi}^{k-1}, \mathbf{y})$.
We will give details about how to get supervisions ${\cal H}^{k}(\bm{\varphi}^{k-1}, \mathbf{y})$ and loss functions $\mathrm{L}^{k}\left(\mathbf{F}, \mathbf{W}^{k}, {\cal H}^{k}(\bm{\varphi}^{k-1}, \mathbf{y})\right)$ in Section~\ref{sec:pcl}.

\begin{algorithm}[t]
{
\caption{{The overall training procedure (one iteration)}}
\label{alg:all}
\begin{algorithmic}[1]
   \REQUIRE An image, its proposal boxes ${\cal B}$, and its image label vector $\mathbf{y} = [y_{1}, ..., y_{C}]^{T}$; refinement times $K$.
   \ENSURE An updated network.
   \STATE Feed the image and ${\cal B}$ into the network to produce proposal score matrices  $\bm{\varphi}^{k}(\mathbf{F}, \mathbf{W}^{k}), k \in \{0, 1, ..., K\}$ (simplified as $\bm{\varphi}^{k}$ later).
   \STATE Compute loss $\mathrm{L}^{0}\left(\mathbf{F}, \mathbf{W}^{0}, \mathbf{y}\right)$ by Eq.~(\ref{equ:loss_base}), see Section~\ref{sec:mil}.
   \FOR{$k=1$ \TO $K$}
      \STATE Generate supervisions ${\cal H}^{k}(\bm{\varphi}^{k-1}, \mathbf{y})$, see Section~\ref{sec:pcl}.
      \STATE Compute loss $\mathrm{L}^{k}\left(\mathbf{F}, \mathbf{W}^{k}, {\cal H}^{k}(\bm{\varphi}^{k-1}, \mathbf{y})\right)$ by Eq.~(\ref{equ:loss_label})/(\ref{equ:loss_label_weighted})/(\ref{equ:loss_bag}), see Section~\ref{sec:pcl}.
   \ENDFOR
   \STATE Optimize $\mathop{\sum} \limits_{k=0}^{K} \mathrm{L}^{k}\left(\mathbf{F}, \mathbf{W}^{k}, {\cal H}^{k}(\bm{\varphi}^{k-1}, \mathbf{y})\right)$, \ie, Eq.~(\ref{equ:loss_all}), \wrt\ $\mathbf{F}, \mathbf{W}^{k}$ (not \wrt\ ${\cal H}^{k}(\bm{\varphi}^{k-1}, \mathbf{y})$).
\end{algorithmic}
}
\end{algorithm}

{During the forward process of each Stochastic Gradient Descent (SGD) training iteration,
we obtain a set of proposal scores of an input image.
Accordingly, we generate the supervisions ${\cal H}^{k}(\bm{\varphi}^{k-1}, \mathbf{y})$ for the iteration to compute the loss Eq.~(\ref{equ:loss_all}).
During the back-propagation process of each SGD training iteration,
we optimize the loss Eq.~(\ref{equ:loss_all}) \wrt\ proposal features $\mathbf{F}$ and classifiers $\mathbf{W}^{k}$.
We summarize this procedure in Algorithm~\ref{alg:all}.}
Note that we do not use an alternating training strategy,
\ie, fixing supervisions and training a complete model,
fixing the model and updating supervisions.
The reasons are that:
1) it is very time-consuming because it requires training models multiple times;
2) training different models in different refinement steps separately may harm the performance because it hinders the process to benefit from the shared proposal features (\ie, $\mathbf{F}$).

\subsection{Proposal cluster learning}
\label{sec:pcl}

Here we will introduce our methods to learn refined instance classifiers based on proposal clusters (\ie, proposal cluster learning).

Recall from Section~\ref{sec:notation} that we have a set of proposals with boxes ${\cal B} = \{b_{r}\}_{r=1}^{R}$.
For the $k$-th refinement,
our goal is to generate supervisions ${\cal H}^{k}(\bm{\varphi}^{k-1}, \mathbf{y})$ for the loss functions $\mathrm{L}^{k}\left(\mathbf{F}, \mathbf{W}^{k}, {\cal H}^{k}(\bm{\varphi}^{k-1}, \mathbf{y})\right)$
using the proposal scores $\bm{\varphi}^{k-1}$ and image label $\mathbf{y}$
in each training iteration.
We use ${\cal H}^{k}, \mathrm{L}^{k}$ later for simplification, dropping the dependence on $\bm{\varphi}^{k-1}, \mathbf{y}, \mathbf{F}, \mathbf{W}^{k}$.

We do this in three steps.
1) We find proposal cluster centers which are proposals corresponding to different objects.
2) We group the remaining proposals into different clusters,
where each cluster is associated with a cluster center or corresponds to the background.
3) We generate the supervisions ${\cal H}^{k}$ for the loss functions $\mathrm{L}^{k}$,
enabling us to train the refined instance classifiers.

For the first step,
we compute proposal cluster centers ${\cal S}^{k} = \{S^{k}_{n}\}_{n=1}^{N^{k}}$
based on $\bm{\varphi}^{k-1}$ and $\mathbf{y}$.
The $n$-th cluster center $S^{k}_{n} = (b^{k}_{n}, y^{k}_{n}, s^{k}_{n})$ is defined in Section~\ref{sec:notation}.
We propose two algorithms to find ${\cal S}^{k}$ in Section~\ref{sec:fpcc}~(1) and (2) (also Algorithm~\ref{alg:h} and Algorithm~\ref{alg:g}),
where the first one was proposed in the conference version paper \cite{Ref:Tang2017multiple}
and the second one is proposed in this paper.

For the second step,
according to the proposal cluster centers ${\cal S}^{k}$,
proposal clusters ${\cal C}^{k} = \{{\cal C}^{k}_{n}\}_{n=1}^{N^{k}+1}$ are generated
(${\cal C}^{k}_{N^{k}+1}$ for background and others for objects).
The $n$-th object cluster ${\cal C}^{k}_{n} = ({\cal B}^{k}_{n}, y^{k}_{n}, s^{k}_{n}), n \neq N^{k}+1$
and the background cluster ${\cal C}^{k}_{n} = ({\cal P}^{k}_{n}, y^{k}_{n}), n = N^{k}+1$
are defined in Section~\ref{sec:notation}.
We use the different notation for the background cluster because background proposals are scattered in each image,
and thus it is hard to determine a cluster center and accordingly a cluster score.
The method to generate ${\cal C}^{k}$ was proposed in the conference version paper and is described in Section~\ref{sec:pcg} (also Algorithm~\ref{alg:pcg}).

For the third step,
supervisions ${\cal H}^{k}$ to train the $k$-th refined instance classifier are generated based on the proposal clusters.
We use two strategies where ${\cal H}^{k}$ are either proposal-level labels indicating whether a proposal belongs to an object class,
or cluster-level labels that treats each proposal cluster as a bag.
Subsequently these are used to compute the loss functions $\mathrm{L}^{k}$.
We propose two approaches to do this as described in Section~\ref{sec:lric}~(1) and (2),
where the first one was proposed in the conference version paper
and the second one is proposed in this paper.

\subsubsection{Finding proposal cluster centers}
\label{sec:fpcc}

In the following we introduce two algorithms to find proposal cluster centers. 

\begin{algorithm}[t]
\caption{Finding proposal cluster centers using the highest scoring proposal}
\label{alg:h}
\begin{algorithmic}[1]
   \REQUIRE Proposal boxes ${\cal B} = \{b_{1}, ..., b_{R}\}$; image label vector $\mathbf{y} = [y_{1}, ..., y_{C}]^{T}$; proposal score matrix $\bm{\varphi}^{k-1}$.
   \ENSURE Proposal cluster centers ${\cal S}^{k}$.
   \STATE Initialize ${\cal S}^{k} = \varnothing$.
   \FOR{$c=1$ \TO $C$}
      \IF{$y_{c} = 1$}
         \STATE Choose the $r^{k}_{c}$-th proposal by Eq.~(\ref{equ:j}).
         \STATE ${\cal S}^{k}\textup{.append}\left((b_{r^{k}_{c}}, c, \varphi^{k-1}_{cr^{k}_{c}})\right)$.
      \ENDIF
   \ENDFOR
\end{algorithmic}
\end{algorithm}

\vspace{0.1cm}
\noindent\textbf{(1) Finding proposal cluster centers using the highest scoring proposal.}
A solution for finding proposal cluster centers is to choose the highest scoring proposal,
as in our conference version paper \cite{Ref:Tang2017multiple}.
As in Algorithm~\ref{alg:h}, suppose an image has object class label $c$ (\ie, $y_{c}=1$).
For the $k$-th refinement,
we first select the $r^{k}_{c}$-th proposal which has the highest score by Eq.~(\ref{equ:j}),
where $\varphi^{k-1}_{cr}$ is the predicted score of the $r$-th proposal,
as defined in Section~\ref{sec:notation}.
\begin{equation}
\label{equ:j}
   r^{k}_{c} = \mathop{\argmax}\limits_{r} \varphi^{k-1}_{cr}.
\end{equation}
Then this proposal is chosen as the cluster center, \ie, $S^{k}_{n} = (b^{k}_{n}, y^{k}_{n}, s^{k}_{n}) = (b_{r^{k}_{c}}, c, \varphi^{k-1}_{cr^{k}_{c}})$,
where $b_{r^{k}_{c}}$ is the box of the $r^{k}_{c}$-th proposal.
$\varphi^{k-1}_{cr}$ is chosen as the confidence score that the $r$-th proposal covers at least part of an object of class $c$,
because $\varphi^{k-1}_{cr}$ is the predicted score of the $r$-th proposal been categorized to class $c$.
Therefore, the highest scoring proposal can probably cover at least part of the object
and thus be chosen as the cluster center.

There is a potential problem that one proposal may be chosen as the cluster centers for multiple object classes.
To avoid this problem,
if one proposal corresponds to the cluster centers for multiple object classes,
this proposal would be chosen as the cluster center only by the class with the highest predicted score
and we re-choose cluster centers for other classes.

\vspace{0.1cm}
\noindent\textbf{(2) Finding proposal cluster centers using graphs of top ranking proposals.}
As stated in Section~\ref{sec:intro},
although we can find good proposal cluster centers using the highest scoring proposal,
this ignores that in natural images there are often more than one object for each category.
Therefore, we propose a new method to find cluster centers using graphs of top ranking proposals.

\begin{algorithm}[t]
\caption{Finding proposal cluster centers using graphs of top ranking proposals}
\label{alg:g}
\begin{algorithmic}[1]
   \REQUIRE Proposal boxes ${\cal B} = \{b_{1}, ..., b_{R}\}$; image label vector $\mathbf{y} = [y_{1}, ..., y_{C}]^{T}$; proposal score matrix $\bm{\varphi}^{k-1}$.
   \ENSURE Proposal cluster centers ${\cal S}^{k}$.
   \STATE Initialize ${\cal S}^{k} = \varnothing$.
   \FOR{$c=1$ \TO $C$}
      \IF{$y_{c} = 1$}
         \STATE Select top ranking proposals with indexes ${\cal R}^{k}_{c}$.
         \STATE Build a graph $G^{k}_{c}$ using the top ranking proposals.
         \REPEAT 
            \STATE Set $r^{k}_{c} = \argmax_{r^{\prime}}\sum_{r \in V^{k}_{c}}e^{k}_{crr^{\prime}}$.
            \STATE Set $s = \max_{r} \varphi^{k-1}_{cr}, r\ \textup{s.t.}\ e^{k}_{crr^{k}_{c}} = 1\ \textup{or}\ r = r^{k}_{c}$.
            \STATE ${\cal S}^{k}\textup{.append}\left((b_{r^{k}_{c}}, c, s)\right)$.
            \STATE Remove the $r$-th proposal box from $V^{k}_{c}$, $\forall r\ \textup{s.t.}\ e^{k}_{crr^{k}_{c}} = 1$ or $r = r^{k}_{c}$.
         \UNTIL{$V^{k}_{c}$ is empty.} 
      \ENDIF
   \ENDFOR
\end{algorithmic}
\end{algorithm}

More specifically, suppose an image has object class label $c$.
We first select the top ranking proposals with indexes ${\cal R}^{k}_{c} = \{r^{k}_{c1}, ..., r^{k}_{cN^{k}_{c}}\}$ for the $k$-th refinement.
Then we build an undirected unweighted graph $G^{k}_{c} = (V^{k}_{c}, E^{k}_{c})$ of these proposals based on spatial similarity,
where vertexes $V^{k}_{c}$ correspond to these top ranking proposals,
and edges $E^{k}_{c} = \{e^{k}_{crr^{\prime}}\} = \{e(v^{k}_{cr}, v^{k}_{cr^{\prime}})\}, r,r^{\prime} \in {\cal R}^{k}_{c}$ correspond to the connections between the vertexes.
$e^{k}_{crr^{\prime}}$ is determined according to the spatial similarity between two vertexes (\ie, proposals) as in Eq.~(\ref{equ:e}),
where $I_{rr^{\prime}}$ is the IoU between the $r$-th and $r^{\prime}$-th proposals and $I_{t}$ is a threshold (\eg, $0.4$).
\begin{equation}
\label{equ:e}
   e_{rr^{\prime}} = 
   \begin{cases}
      1 & \textup{if} \ I_{rr^{\prime}} > I_{t},\\
      0 & \textup{otherwise.}
   \end{cases}
\end{equation}
Therefore, two vertexes are connected if they are spatially adjacent.
After that, we greedily generate some cluster centers for class $c$ using this graph.
That is, we iteratively select vertexes which have most connections to be the cluster centers, as in Algorithm~\ref{alg:g}.
{The number of cluster centers (\ie, $N^{k}$) changes for each image in each training iteration
because the top ranking proposals ${\cal R}^{k}_{c}$ change.
See} {Section~\ref{sec:nk} for some typical values of $N^{k}$.}
We use the same method as in Section~\ref{sec:fpcc}~(1)
to avoid one proposal been chosen as the cluster centers for multiple object classes.

The reasons for this strategy are as follows.
First, according to our observation,
the top ranking proposals can always cover at least parts of objects,
thus generating centers from these proposals encourages the selected centers to meet our requirements.
Second, because these proposals cover objects well, better proposals (covering more parts of objects) should have more spatially overlapped proposals (\ie, have more connections).
Third, these centers are spatially far apart,
and thus different centers can correspond to different objects.
This method also has the attractive characteristic that it can generate adaptive number of proposals for each object class,
which is desirable because in natural images there are arbitrary number of objects per-class.
We set the score of the $n$-th proposal cluster center $s^{k}_{n}$ by
$$s^{k}_{n} = \max_{r} \varphi^{k-1}_{cr}, r\ \textup{s.t.}\ e^{k}_{crr^{k}_{c}} = 1\ \textup{or}\ r = r^{k}_{c}$$
(see the $8$-th line in Algorithm~\ref{alg:g})
because if the adjacent proposals of a center proposal have high confidence to cover at least part of an object
(\ie, have high classification scores)
the center proposal should also have such high confidence.

There is an important issue for the graph-based method: how to select the top ranking proposals?
{A simple method is to select proposals whose scores exceed a threshold.
But in our case, proposal scores change in each training iteration,
and thus it is hard to determine a threshold.
Instead,}
for each positive object class,
we use the $k$-means \cite{Ref:Macqueen1967} algorithm to divide proposal scores of an image into some clusters,
and choose proposals in the cluster which has the highest score center to form the top ranking proposals.
{This method ensures that we can select the top ranking proposals although proposal scores change during training.}
Other choices are possible,
but this method works well in experiments.

\begin{algorithm}[t]
\caption{Generating proposal clusters}
\label{alg:pcg}
\begin{algorithmic}[1]
    \REQUIRE Proposal boxes ${\cal B} = \{b_{1}, ..., b_{R}\}$; proposal cluster centers ${\cal S}^{k} = \{S^{k}_{1}, ..., S^{k}_{N^{k}}\}$.
    \ENSURE Proposal clusters ${\cal C}^{k}$.
    \STATE Initialize ${\cal B}^{k}_{n} = \varnothing, \forall n \neq N^{k}+1$.
    \STATE Set $y^{k}_{n}, s^{k}_{n}$ of ${\cal C}^{k}_{n}$ to $y^{k}_{n}, s^{k}_{n}$ of $S^{k}_{n}$, $\forall n \neq N^{k}+1$.
    \STATE Initialize ${\cal P}^{k}_{N^{k}+1} = \varnothing$ and set $y^{k}_{N^{k}+1} = C + 1$.
    \FOR{$r=1$ \TO $R$}
      \STATE Compute IoUs $\{I^{k}_{r1}, ..., I^{k}_{rN^{k}}\}$.
      \STATE Choose the most spatially adjacent center $S^{k}_{n^{k}_{r}}$.
      \IF{$I^{k}_{rn^{k}_{r}} > I^{\prime}_{t}$}
        \STATE ${\cal B}^{k}_{n^{k}_{r}}\textup{.append}\left(b_{r}\right)$.
      \ELSE
        \STATE ${\cal P}^{k}_{N^{k}+1}\textup{.append}\left((b_{r}, s^{k}_{n^{k}_{r}})\right)$.
      \ENDIF
    \ENDFOR
\end{algorithmic}
\end{algorithm}

\subsubsection{Generating proposal clusters}
\label{sec:pcg}

After the cluster centers are found,
we generate the proposal clusters as in our conference version paper \cite{Ref:Tang2017multiple}.
Except for the cluster for background,
good proposal clusters require that proposals in the same cluster are associated with the same object,
and thus proposals in the same cluster should be spatially adjacent.
Specially, given the $r$-th proposal,
we compute a set of IoUs $\{I^{k}_{r1}, ..., I^{k}_{rN^{k}}\}$,
where $I^{k}_{rn}$ is the IoU between the $r$-th proposal and the box $b^{k}_{n}$ of the $n$-th cluster center.
Then we assign the $r$-th proposal to the $n^{k}_{r}$-th object cluster if $I^{k}_{rn^{k}_{r}}$ is larger than a threshold $I^{\prime}_{t}$ (\eg, $0.5$)
and to the background cluster otherwise,
where $n^{k}_{r}$ is the index of the most spatially adjacent cluster center as Eq.~(\ref{equ:nkr}).
\begin{equation}
\label{equ:nkr}
    n^{k}_{r} = \mathop{\argmax}\limits_{n} I^{k}_{rn}.
\end{equation}
The overall procedures to generate proposal clusters are summarized in Algorithm~\ref{alg:pcg}.
We set the proposal scores for the background cluster to the scores of their most spatially adjacent centers as the 10-the line in Algorithm~\ref{alg:pcg},
because if the cluster center $S^{k}_{n}$ has confidence $s^{k}_{n}$ that it covers an object,
the proposal far away from $S^{k}_{n}$ should have confidence $s^{k}_{n}$ to be background.

\subsubsection{Learning refined instance classifiers}
\label{sec:lric}

To get supervisions ${\cal H}^{k}$ and loss functions $\mathrm{L}^{k}$ to learn the $k$-th refined instance classifier,
we design two approaches as follows.

\vspace{0.1cm}
\noindent\textbf{(1) Assigning proposals object labels.}
The most straightforward way to refine classifiers is to directly assign object labels to all proposals in object clusters because these proposals potentially correspond to whole objects,
as in our conference version paper \cite{Ref:Tang2017multiple}.
As the cluster centers covers at least parts of objects,
their adjacent proposals (\ie, proposals in the cluster) can contain larger parts of objects.
Accordingly, we can assign the cluster label $y^{k}_{n}$ to all proposals in the $n$-th cluster.

More specifically, the supervisions ${\cal H}^{k}$ are proposal-level labels,
\ie, ${\cal H}^{k} = \{\mathbf{y}^{k}_{r}\}_{r=1}^{R}$.
$\mathbf{y}^{k}_{r} = [y^{k}_{1r}, ..., y^{k}_{(C+1)r}]^{T} \in \mathbb{R}^{(C+1) \times 1}$
is the label vector of the $r$-th proposal for the $k$-th refinement,
where $y^{k}_{y^{k}_{n}r} = 1$ and $y^{k}_{cr} = 0, c \neq y^{k}_{n}$ if the $r$-th proposal belongs to the $n$-th clusters.
Consequently, we use the standard softmax loss function to train the refined classifiers as in Eq.~(\ref{equ:loss_label}),
where $\varphi^{k}_{cr}$ is the predicted score of the $r$-th proposal as defined in Section~\ref{sec:notation}.
\begin{equation}
\label{equ:loss_label}
   \mathrm{L}^{k}\left(\mathbf{F}, \mathbf{W}^{k}, {\cal H}^{k}\right) = -\frac{1}{R}\mathop{\sum} \limits_{r=1}^{R} \sum_{c=1}^{C+1} y^{k}_{cr} \log \varphi^{k}_{cr}.
\end{equation}
Through iterative instance classifier refinement (\ie, multiple times of refinement as $k$ increase),
the detector detects larger parts of objects gradually by forcing the network to ``see'' larger parts of objects.

Actually, the so learnt supervisions ${\cal H}^{k}$ are very noisy,
especially in the beginning of training.
This results in unstable solutions.
To solve this problem, we change the loss in Eq.~(\ref{equ:loss_label}) to a weighted version, as in Eq.~(\ref{equ:loss_label_weighted}).
\begin{equation}
\label{equ:loss_label_weighted}
   \mathrm{L}^{k}\left(\mathbf{F}, \mathbf{W}^{k}, {\cal H}^{k}\right) = -\frac{1}{R}\mathop{\sum} \limits_{r=1}^{R} \sum_{c=1}^{C+1} \lambda^{k}_{r} y^{k}_{cr} \log \varphi^{k}_{cr}.
\end{equation}
$\lambda^{k}_{r}$ is the loss weight that is the same as the cluster confidence score $s^{k}_{n}$ for object clusters or proposal confidence score $s^{k}_{nm}$ for the background cluster
if the $r$-th proposal belongs to the $n$-th cluster.
From Algorithm~\ref{alg:pcg}, we can observe that $\lambda^{k}_{r}$ is the same as the cluster center confidence score $s^{k}_{n}$.
The reasons for this strategy are as follows.
In the beginning of training,
although we cannot obtain good proposal clusters,
each $s^{k}_{n}$ is small, hence each $\lambda^{k}_{r}$ is small and the loss is also small.
As a consequence, the performance of the network will not decrease a lot.
During the training, the top ranking proposals will cover objects well,
and thus we can generate good proposal clusters.
Then we can train satisfactory instance classifiers.

\vspace{0.1cm}
\noindent\textbf{(2) Treating clusters as bags.}
As we stressed before,
although directly assigning proposals object labels can boost the results,
it may confuse the network because we simultaneously assign the same label to different parts of objects.
Focusing on this,
we further propose to treat each proposal cluster as a small new bag and use the cluster label as the bag label.
Thus the supervisions ${\cal H}^{k}$ for the $k$-th refinement are bag-level (cluster-level) labels,
\ie, ${\cal H}^{k} = \{y^{k}_{n}\}_{n=1}^{N^{k}+1}$.
$y^{k}_{n}$ is the label of the $n$-th bag, \ie, the label of the $n$-th proposal cluster,
as defined in Section~\ref{sec:notation}.

Specially, for object clusters,
we choose average MIL pooling,
because these proposals should cover at least parts of objects and thus should have relatively high prediction scores.
For the background cluster,
we assign the background label to all proposals in the cluster according to the MIL constraints
(all instances in negative bags are negative).
Then the loss function for refinement will be Eq.~(\ref{equ:loss_bag}).
\begin{equation}
\label{equ:loss_bag}
\begin{aligned}
   \mathrm{L}^{k}\left(\mathbf{F}, \mathbf{W}^{k}, {\cal H}^{k}\right) = -\frac{1}{R} (&\mathop{\sum} \limits_{n=1}^{N^{k}} s^{k}_{n} M_{n}^{k} \log \frac{\mathop{\sum} \limits_{r \ \textup{s.t.}\ b_{r} \in {\cal B}^{k}_{n}} \varphi^{k}_{y^{k}_{n}r}}{M_{n}^{k}}\\
   + &\mathop{\sum} \limits_{r \in {\cal C}^{k}_{N^{k} + 1}} \lambda^{k}_{r} \log \varphi^{k}_{(C+1)r}).
\end{aligned}
\end{equation}
$s^{k}_{n}$, $M_{n}^{k}$, and $\varphi^{k}_{cr}$ are the cluster confidence score of the $n$-th object cluster,
the number of proposals in the $n$-th cluster,
and the predicted score of the $r$-th proposal,
respectively,
as defined in Section~\ref{sec:notation}.
$b_{r} \in {\cal B}^{k}_{n}$ and $r \in {\cal C}^{k}_{N^{k}+1}$ indicate that
the $r$-th proposal belongs to the $n$-th object cluster
and the background cluster respectively.

Compared with the directly assigning label approach,
this method tolerates some proposals to have low scores, which can reduce the ambiguities to some extent.

\subsection{Testing}

During testing,
the proposal scores of refined instance classifiers are used as the final detection scores,
as the blue arrows in Fig.~\ref{fig:architecture}.
Here the mean output of all refined classifiers is chosen.
The Non-Maxima Suppression (NMS) is used to filter out redundant detections.

\section{Experiments}
\label{sec:exp}

In this section, we first introduce our experimental setup including datasets, evaluation metrics, and implementation details.
Then we conduct elaborate experiments to discuss the influence of different settings.
Next, we compare our results with others to show the effectiveness of our method.
After that, we show some qualitative results for further analyses.
{Finally, we give some runtime analyses of our method.
Codes for reproducing our results are available at \url{https://github.com/ppengtang/oicr/tree/pcl}.}

\subsection{Experimental setup}
\label{sec:exp_setup}

\subsubsection{Datasets and evaluation metrics}
We evaluate our method on four challenging datasets: the PASCAL VOC 2007 and 2012 datasets \cite{Ref:Everingham2015}, the ImageNet detection dataset \cite{Ref:Russakovsky2015}, {and the MS-COCO dataset \cite{Ref:Lin2014}}.
Only image-level annotations are used to train our models.

The PASCAL VOC 2007 and 2012 datasets have $9,962$ and $22,531$ images respectively for $20$ object classes.
These two datasets are divided into train, val, and test sets.
Here we choose the trainval set ($5,011$ images for 2007 and $11,540$ images for 2012) to train our network.
For testing, there are two metrics for evaluation: mAP and CorLoc.
Following the standard PASCAL VOC protocol \cite{Ref:Everingham2015},
Average Precision (AP) and the mean of AP (mAP) is the evaluation metric to test our model on the testing set.
Correct Localization (CorLoc) is to test our model on the training set measuring the localization accuracy \cite{Ref:Deselaers2012}.
All these two metrics are based on the PASCAL criterion, \ie, IoU$>$0.5 between groundtruth boundingboxes and predicted boxes.

The ImageNet detection dataset has hundreds of thousands of images with $200$ object classes.
It is also divided into train, val, and test sets.
Following \cite{Ref:Girshick2016}, we split the val set into val1 and val2, and randomly choose at most $1$K images in the train set for each object class (we call it train$_{\textup{1K}}$).
We train our model on the mixture of train$_{\textup{1K}}$ and val1 sets, and test it on the val2 set,
which will lead to $160,651$ images for training and $9,916$ images for testing.
We also use the mAP for evaluation on the ImageNet.

{
The MS-COCO dataset has $80$ object classes
and is divided into train, val, and test sets.
Since the groundtruths on the test set are not released,
we train our model on the MS-COCO 2014 train set (about $80$K images) and test it on the val set (about $40$K images).
For evaluation, we use two metrics mAP@0.5 and mAP@[.5, .95]
which are the standard PASCAL criterion (\ie, IoU$>$0.5)
and the standard MS-COCO criterion
(\ie, computing the average of mAP for IoU$\in$[0.5 : 0.05 : 0.95]) respectively.
}

\subsubsection{Implementation details}
Our method is built on two pre-trained ImageNet \cite{Ref:Russakovsky2015} networks VGG$\_$M \cite{Ref:Chatfield2014} and VGG16 \cite{Ref:Simonyan2015},
each of which has some conv layers with max-pooling layers and three fc layers.
We replace the last max-pooling layer by the SPP layer,
and the last fc layer as well as the softmax loss layer by the layers described in Section~\ref{sec:method}.
To increase the feature map size from the last conv layer,
we replace the penultimate max-pooling layer and its subsequent conv layers by the dilated conv layers \cite{Ref:Yu2016,Ref:Chen2017}.
The newly added layers are initialized using Gaussian distributions with $0$-mean and standard deviations $0.01$.
Biases are initialized to $0$.

During training, the mini-batch size for SGD is set to be $2$, $32$, {and $4$} for PASCAL VOC, ImageNet, {and MS-COCO}, respectively.
The learning rate is set to $0.001$ for the first $40$K, $60$K, $15$K, {and $85$K} iterations for the PASCAL VOC 2007, PASCAL VOC 2012, ImageNet, {and MS-COCO} datasets, respectively.
Then we decrease the learning rate to $0.0001$ in the following $10$K, $20$K, $5$K, {and $20$K} iterations for the PASCAL VOC 2007, PASCAL VOC 2012, ImageNet, {and MS-COCO} datasets, respectively.
The momentum and weight decay are set to be $0.9$ and $0.0005$ respectively.

Selective Search \cite{Ref:Uijlings2013}, EdgeBox \cite{Ref:Zitnick2014}, and MCG \cite{Ref:Pont2017} are adopted to generate about $2,000$ proposals per-image for the PASCAL VOC, ImageNet, and MS-COCO datasets, respectively.
For data augmentation, we use five image scales $\{480, 576, 688, 864, 1200\}$ (resize the shortest side to one of these scales) with horizontal flips for both training and testing.
If not specified,
the instance classifiers are refined three times,
\ie, $K=3$ in Section~\ref{sec:ots},
so there are four output streams;
the IoU threshold $I_{t}$ in Section~\ref{sec:fpcc}~(2) (also Eq.~(\ref{equ:e})) is set to $0.4$;
the number of $k$-means clusters in the last paragraph of Section~\ref{sec:fpcc}~(2) is set to $3$;
$I^{\prime}_{t}$ in Section~\ref{sec:pcg} (also the $5$-th line of Algorithm~\ref{alg:pcg}) is set to $0.5$.

Similar to other works \cite{Ref:Li2016,Ref:Diba2017,Ref:Jie2017},
we train a supervised object detector through choosing the top-scoring proposals given by our method as pseudo groundtruths to further improve our results.
Here we train a Fast R-CNN (FRCNN) \cite{Ref:Girshick2015} using the VGG16 model and the same five image scales (horizontal flips only in training).
The same proposals are chosen to train and test the FRCNN.
NMS (with $30\%$ IoU threshold) is applied to compute AP.

Our experiments are implemented based on the Caffe \cite{Ref:Jia2014} deep learning framework, using Python and C++.
The $k$-means algorithm to produce top ranking proposals is implemented by scikit-learn \cite{Ref:Pedregosa2011}.
All of our experiments are running on an NVIDIA GTX TitanX Pascal GPU
and Intel(R) i7-6850K CPU (3.60GHz).

\subsection{Discussions}
\label{sec:abl_exp}

We first conduct some experiments
to discuss the influence of different components of our method
(including instance classifier refinement,
different proposal generation methods,
different refinement strategies,
and weighted loss)
and different parameter settings
(including the IoU threshold $I_{t}$ defined in Section~\ref{sec:fpcc}~(2),
the number of $k$-means clusters described in Section~\ref{sec:fpcc}~(2),
the IoU threshold $I^{\prime}_{t}$ defined in Section~\ref{sec:pcg},
{and multi-scale training and testing}.)
{We also discuss the number of proposal cluster centers.}
Without loss of generality, we only perform experiments on the VOC 2007 dataset and use the VGG$\_$M model.

\subsubsection{The influence of instance classifier refinement}
\label{sec:influence_lp}

\begin{figure}[t]
\begin{center}
   \includegraphics[width=\linewidth]{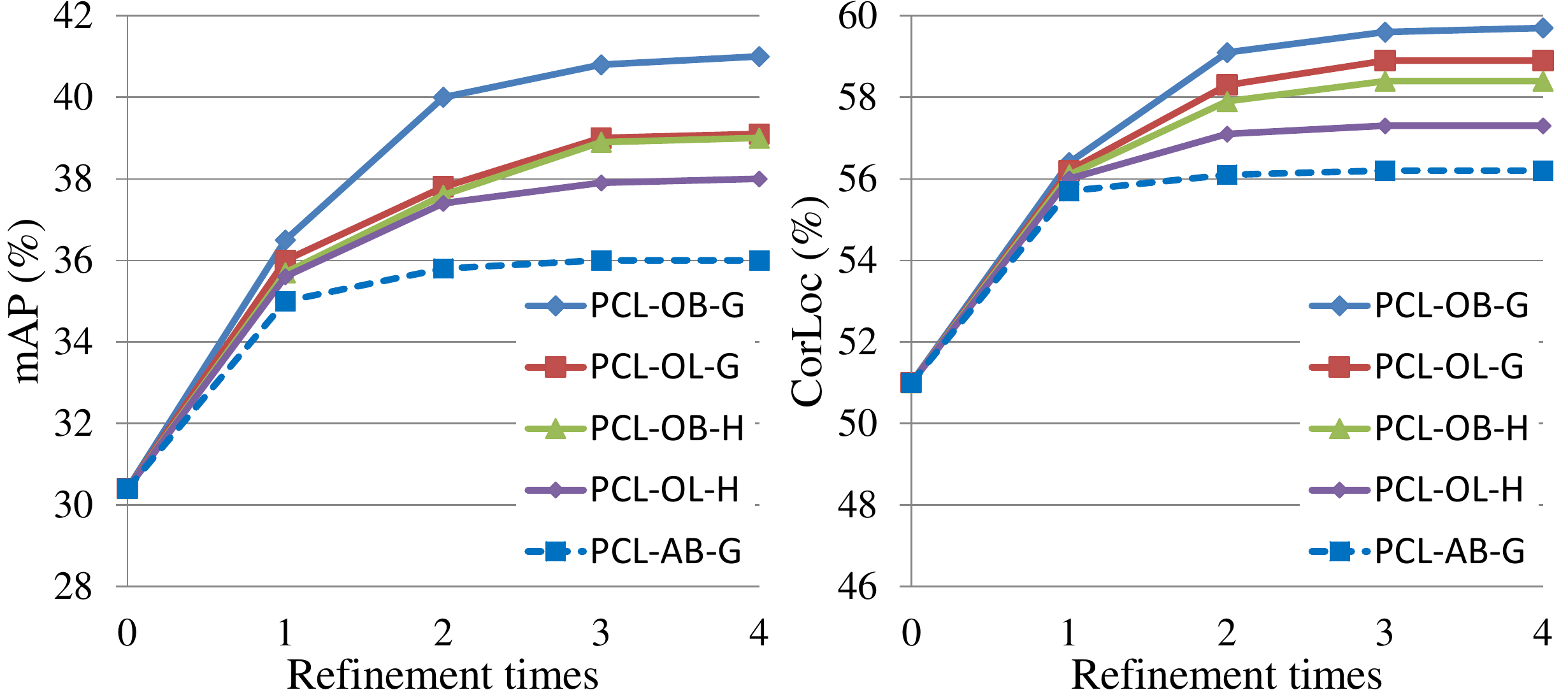}
\end{center}
   \caption{Results on VOC 2007 for different refinement times and different training strategies,
   where ``\methodname-xx-H'' and ``\methodname-xx-G'' indicate the highest scoring proposal based method and the graph-based method to generate proposal clusters respectively,
   ``\methodname-OL-x'' and ``\methodname-OB-x'' indicate the directly assigning label method and the treating clusters as bags method to train the network online respectively,
   and ``\methodname-AB-x'' indicates using the alternating training strategy.}
\label{fig:ablation_refinement_time}
\end{figure}

As the five curves in Fig.~\ref{fig:ablation_refinement_time} show,
we observe that compared with the basic MIL network,
for both refinement methods,
even refining instance classifier a single time boosts the performance a lot.
This confirms the necessity of refinement.
If we refine the classifier multiple times, the results are improved further.
But when refinement is implemented too many times,
the performance gets saturated (there are no obvious improvements from $3$ times to $4$ times).
This is because the network tends to converge so that the supervision of the $4$-th time is similar to the $3$-rd time.
In the rest of this paper we only refine classifiers $3$ times.
Notice that in Fig.~\ref{fig:ablation_refinement_time},
the ``0 time'' is similar to the WSDDN \cite{Ref:Bilen2016} using Selective Search as proposals.

\subsubsection{The influence of different proposal cluster generation methods}

We discuss the influence of different proposal cluster generation methods.
As shown in the Fig.~\ref{fig:ablation_refinement_time}
({green and purple solid curves for the highest scoring proposal based method, blue and red solid curves for the graph-based method}),
for all refinement times,
the graph-based method obtains better performance, because it can generate better cluster centers.
Thus we choose the graph-based method in the rest of our paper.

\subsubsection{The influence of different refinement strategies}

We then show the influence of different refinement strategies.
The directly assigning label method is replaced by treating clusters as bags ({blue and green solid curves}).
From Fig.~\ref{fig:ablation_refinement_time}, it is obvious that the results by treating clusters as bags are better.
In addition, compared with the alternating training strategy (blue dashed curve),
our online training boosts the performance consistently and significantly, which confirms the necessity of sharing proposal features.
Online training also reduces the training time a lot,
because it only requires training a single model instead of training $K+1$ models for $K$ times refinement in the alternating strategy.
In the rest of our paper,
we only report results by the ``\methodname-OB-G'' method in Fig.~\ref{fig:ablation_refinement_time}
because it achieves the best performance.

\subsubsection{The influence of weighted loss}
\label{sec:influence_wl}

We also study the influence of our weighted loss in Eq.~(\ref{equ:loss_bag}).
Note that Eq.~(\ref{equ:loss_bag}) can be easily changed to the unweighted version by simply setting $\lambda^{k}_{r}$ and $s^{k}_{n}$ to be $1$.
Here we train a network using the unweighted loss.
The results of the unweighted loss are mAP $33.6\%$ and CorLoc $51.2\%$.
We see that if we use the unweighted loss, the improvement from refinement is very scant and the performance is even worse than the alternating strategy.
Using the weighted loss achieves much better performance (mAP $40.8\%$ and CorLoc $59.6\%$),
which confirms our theory in Section~\ref{sec:lric}.

\begin{figure}[t]
\begin{minipage}[t]{0.485\linewidth}
    \begin{center}
        \includegraphics[width=\linewidth]{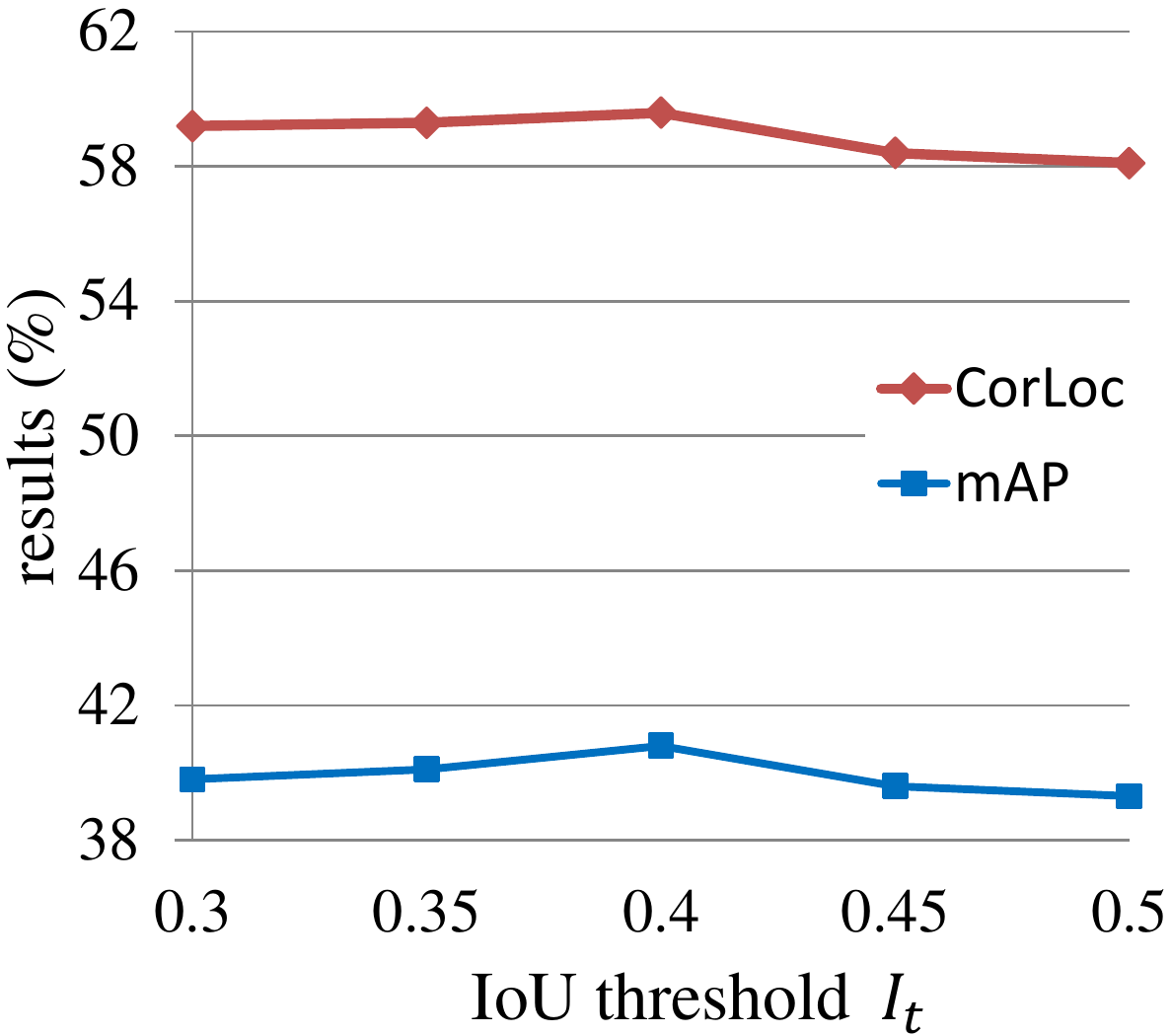}
    \end{center}
    \caption{Results on VOC 2007 for different IoU threshold $I_{t}$.}
    \label{fig:ablation_threshold}
\end{minipage}\hfill
\begin{minipage}[t]{0.485\linewidth}
    \begin{center}
        \includegraphics[width=\linewidth]{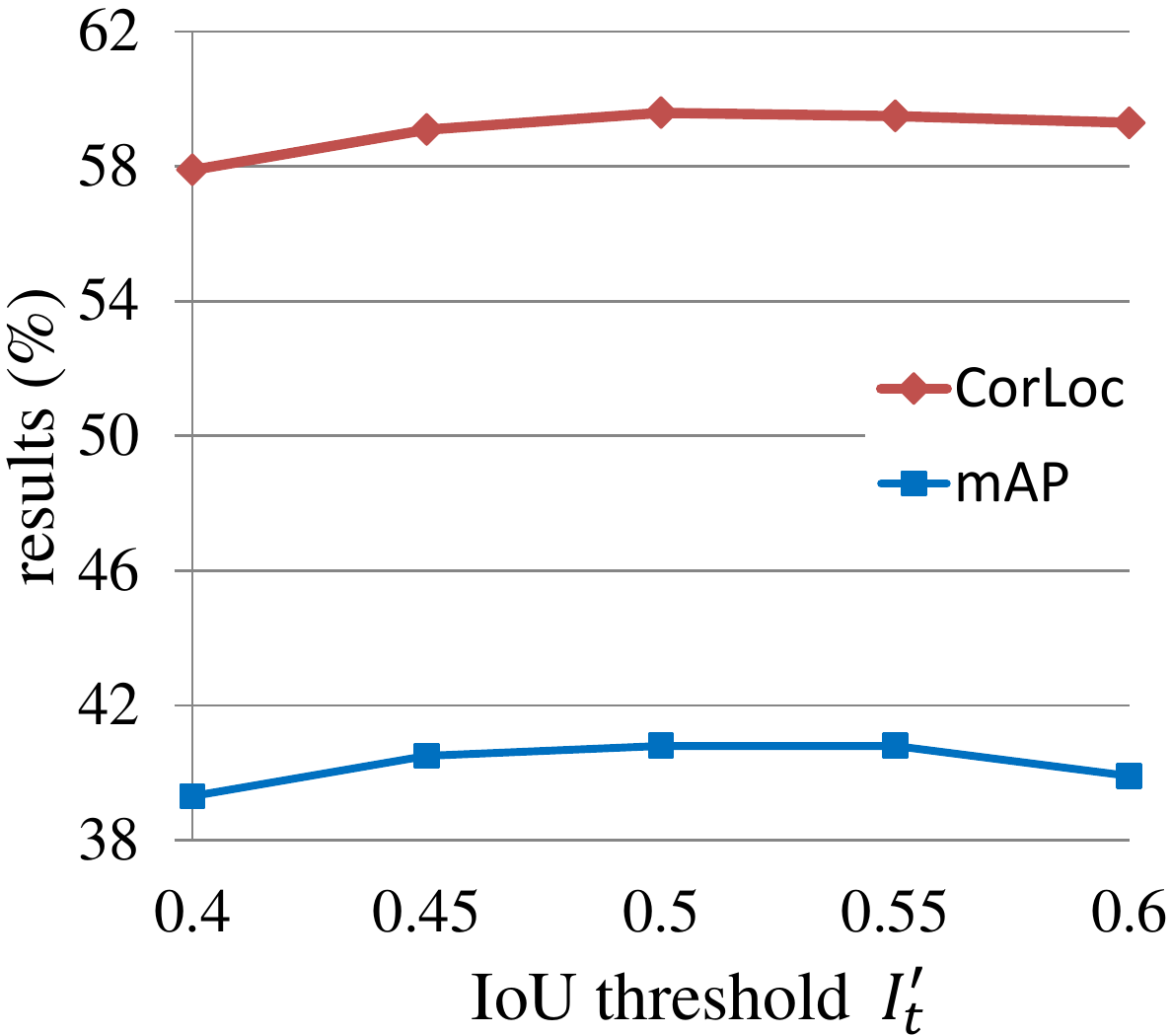}
    \end{center}
    \caption{Results on VOC 2007 for different IoU threshold $I^{\prime}_{t}$.}
    \label{fig:ablation_threshold1}
\end{minipage}
\end{figure}

{\addtolength{\tabcolsep}{-2pt}
\begin{table*}[t]
\caption{Results (AP in $\%$) for different methods on the VOC 2007 test set.
The upper part shows the results using a single model.
The lower part shows the results of combing multiple models.
See Section~\ref{sec:compar} for the definitions of the \methodname-based methods.}
\begin{center}
\resizebox{\linewidth}{!}{
\begin{tabular}{|@{}L{3.15cm}|*{20}{x}|x|}
  \hline
  Method & aero & bike & bird & boat & bottle & bus & car & cat & chair & cow & table & dog & horse & mbike & person & plant & sheep & sofa & train & tv & mAP \\
  \hline\hline
  WSDDN-VGG$\_$F \cite{Ref:Bilen2016} & 42.9 & 56.0 & 32.0 & 17.6 & 10.2 & 61.8 & 50.2 & 29.0 & 3.8 & 36.2 & 18.5 & 31.1 & 45.8 & 54.5 & 10.2 & 15.4 & 36.3 & 45.2 & 50.1 & 43.8 & 34.5\\
  WSDDN-VGG$\_$M \cite{Ref:Bilen2016} & 43.6 & 50.4 & 32.2 & $\bm{26.0}$ &  9.8 & 58.5 & 50.4 & 30.9 & 7.9 & 36.1 & 18.2 & 31.7 & 41.4 & 52.6 & 8.8 & 14.0 & 37.8 & 46.9 & 53.4 & 47.9 & 34.9\\
  WSDDN-VGG16 \cite{Ref:Bilen2016} & 39.4 & 50.1 & 31.5 & 16.3 & 12.6 & $\bm{64.5}$ & 42.8 & 42.6 & 10.1 & 35.7 & 24.9 & 38.2 & 34.4 & 55.6 & 9.4 & 14.7 & 30.2 & 40.7 & 54.7 & 46.9 & 34.8\\
  WSDDN+context \cite{Ref:Kantorov2016} & $\bm{57.1}$ & 52.0 & 31.5 & 7.6 & 11.5 & 55.0 & 53.1 & 34.1 & 1.7 & 33.1 & $\bm{49.2}$ & $\bm{42.0}$ & 47.3 & 56.6 & 15.3 & 12.8 & 24.8 & 48.9 & 44.4 & 47.8 & 36.3\\
  \hline
  \methodname-OB-G-VGG$\_$M & 54.0 & 60.8 & 33.9 & 18.4 & $\bm{15.8}$ & 57.7 & 59.8 & $\bm{52.0}$ & 2.7 & 48.3 & 46.4 & 38.5 & $\bm{47.9}$ & 63.9 & 7.0 & 21.7 & 38.1 & 42.1 & 54.3 & 53.0 & 40.8\\
  \methodname-OB-G-VGG16 & 54.4 & $\bm{69.0}$ & $\bm{39.3}$ & 19.2 & 15.7 & 62.9 & $\bm{64.4}$ & 30.0 & \underline{$\bm{25.1}$} & $\bm{52.5}$ & 44.4 & 19.6 & 39.3 & $\bm{67.7}$ & \underline{$\bm{17.8}$} & $\bm{22.9}$ & $\bm{46.6}$ & \underline{$\bm{57.5}$} & $\bm{58.6}$ & \underline{$\bm{63.0}$} & $\bm{43.5}$\\
  \hline\hline
  WSDDN-Ens. \cite{Ref:Bilen2016} & 46.4 & 58.3 & 35.5 & 25.9 & 14.0 & 66.7 & 53.0 & 39.2 & 8.9 & 41.8 & 26.6 & 38.6 & 44.7 & 59.0 & 10.8 & 17.3 & 40.7 & 49.6 & 56.9 & 50.8 & 39.3 \\
  OM+MIL+FRCNN \cite{Ref:Li2016} & 54.5 & 47.4 & 41.3 & 20.8 & 17.7 & 51.9 & 63.5 & 46.1 & $\bm{21.8}$ & 57.1 & 22.1 & 34.4 & 50.5 & 61.8 & 16.2 & \underline{$\bm{29.9}$} & 40.7 & 15.9 & 55.3 & 40.2 & 39.5\\
  WCCN \cite{Ref:Diba2017} & 49.5 & 60.6 & 38.6 & \underline{$\bm{29.2}$} & 16.2 & 70.8 & 56.9 & 42.5 & 10.9 & 44.1 & 29.9 & \underline{$\bm{42.2}$} & 47.9 & 64.1 & 13.8 & 23.5 & 45.9 & 54.1 & 60.8 & 54.5 & 42.8\\
  Jie~\etal\ \cite{Ref:Jie2017} & 54.2 & 52.0 & 35.2 & 25.9 & 15.0 & 59.6 & 67.9 & \underline{$\bm{58.7}$} & 10.1 & \underline{$\bm{67.4}$} & 27.3 & 37.8 & 54.8 & 67.3 & 5.1 & 19.7 & \underline{$\bm{52.6}$} & 43.5 & 56.9 & $\bm{62.5}$ & 43.7\\
  \hline
  \methodname-OB-G-Ens. & 57.1 & 67.1 & 40.9 & 16.9 & 18.8 & 65.1 & 63.7 & 45.3 & 17.0 & 56.7 & 48.9 & 33.2 & 54.4 & \underline{$\bm{68.3}$} & $\bm{16.8}$ & 25.7 & 45.8 & 52.2 & 59.1 & 62.0 & 45.8\\
  \methodname-OB-G-Ens.+FRCNN & \underline{$\bm{63.2}$} & \underline{$\bm{69.9}$} & \underline{$\bm{47.9}$} & 22.6 & \underline{$\bm{27.3}$} & \underline{$\bm{71.0}$} & \underline{$\bm{69.1}$} & 49.6 & 12.0 & 60.1 & \underline{$\bm{51.5}$} & 37.3 & \underline{$\bm{63.3}$} & 63.9 & 15.8 & 23.6 & 48.8 & $\bm{55.3}$ & \underline{$\bm{61.2}$} & 62.1 & \underline{$\bm{48.8}$}\\
\hline
\end{tabular}
}
\end{center}
\label{table:voc_2007_map}
\end{table*}
}

{\addtolength{\tabcolsep}{-2pt}
\begin{table*}[t]
\caption{Results (CorLoc in $\%$) for different methods on the VOC 2007 trainval set.
The upper part shows the results using a single model.
The lower part shows the results of combing multiple models.
See Section~\ref{sec:compar} for the definitions of the \methodname-based methods.}
\begin{center}
\resizebox{\linewidth}{!}{
\begin{tabular}{|@{}L{3.15cm}|*{20}{x}|x|}
  \hline
  Method & aero & bike & bird & boat & bottle & bus & car & cat & chair & cow & table & dog & horse & mbike & person & plant & sheep & sofa & train & tv & mean \\
  \hline\hline
  WSDDN-VGG$\_$F \cite{Ref:Bilen2016} & 68.5 & 67.5 & 56.7 & 34.3 & 32.8 & 69.9 & 75.0 & 45.7 & 17.1 & 68.1 & 30.5 & 40.6 & 67.2 & 82.9 & 28.8 & 43.7 & 71.9 & 62.0 & 62.8 & 58.2 & 54.2\\
  WSDDN-VGG$\_$M \cite{Ref:Bilen2016} & 65.1 & 63.4 & 59.7 & 45.9 & 38.5 & 69.4 & 77.0 & 50.7 & 30.1 & 68.8 & 34.0 & 37.3 & 61.0 & 82.9 & 25.1 & 42.9 & \underline{$\bm{79.2}$} & 59.4 & 68.2 & 64.1 & 56.1\\
  WSDDN-VGG16 \cite{Ref:Bilen2016} & 65.1 & 58.8 & 58.5 & 33.1 & $\bm{39.8}$ & 68.3 & 60.2 & 59.6 & 34.8 & 64.5 & 30.5 & 43.0 & 56.8 & 82.4 & 25.5 & 41.6 & 61.5 & 55.9 & 65.9 & 63.7 & 53.5\\
  WSDDN+context \cite{Ref:Kantorov2016} & $\bm{83.3}$ & 68.6 & 54.7 & 23.4 & 18.3 & 73.6 & 74.1 & 54.1 & 8.6 & 65.1 & 47.1 & \underline{$\bm{59.5}$} & 67.0 & 83.5 & \underline{$\bm{35.3}$} & 39.9 & 67.0 & 49.7 & 63.5 & 65.2 & 55.1\\
  \hline
  \methodname-OB-G-VGG$\_$M & 78.7 & 75.7 & 55.9 & 33.5 & 35.9 & 72.6 & 81.2 & \underline{$\bm{69.5}$} & 10.1 & 74.7 & $\bm{52.5}$ & 55.1 & $\bm{73.1}$ & 87.6 & 15.9 & 46.2 & 70.1 & 60.5 & 71.9 & 71.3 & 59.6\\
  \methodname-OB-G-VGG16 & 79.6 & \underline{$\bm{85.5}$} & $\bm{62.2}$ & \underline{$\bm{47.9}$} & 37.0 & \underline{$\bm{83.8}$} & $\bm{83.4}$ & 43.0 & \underline{$\bm{38.3}$} & $\bm{80.1}$ & 50.6 & 30.9 & 57.8 & $\bm{90.8}$ & 27.0 & \underline{$\bm{58.2}$} & 75.3 & \underline{$\bm{68.5}$} & $\bm{75.7}$ & $\bm{78.9}$ & $\bm{62.7}$\\
  \hline\hline
  OM+MIL+FRCNN \cite{Ref:Li2016} & 78.2 & 67.1 & 61.8 & 38.1 & 36.1 & 61.8 & 78.8 & 55.2 & 28.5 & 68.8 & 18.5 & 49.2 & 64.1 & 73.5 & 21.4 & 47.4 & 64.6 & 22.3 & 60.9 & 52.3 & 52.4\\
  WSDDN-Ens. \cite{Ref:Bilen2016} & 68.9 & 68.7 & 65.2 & 42.5 & 40.6 & 72.6 & 75.2 & 53.7 & 29.7 & 68.1 & 33.5 & 45.6 & 65.9 & 86.1 & 27.5 & 44.9 & $\bm{76.0}$ & 62.4 & 66.3 & 66.8 & 58.0\\
  WCCN \cite{Ref:Diba2017} & \underline{$\bm{83.9}$} & 72.8 & 64.5 & $\bm{44.1}$ & 40.1 & 65.7 & 82.5 & 58.9 & $\bm{33.7}$ & 72.5 & 25.6 & 53.7 & 67.4 & 77.4 & 26.8 & 49.1 & 68.1 & 27.9 & 64.5 & 55.7 & 56.7\\
  Jie \etal\ \cite{Ref:Jie2017} & 72.7 & 55.3 & 53.0 & 27.8 & 35.2 & 68.6 & 81.9 & $\bm{60.7}$ & 11.6 & 71.6 & 29.7 & $\bm{54.3}$ & 64.3 & 88.2 & 22.2 & 53.7 & 72.2 & 52.6 & 68.9 & 75.5 & 56.1\\
  \hline
  \methodname-OB-G-Ens. & 81.7 & 82.4 & 63.4 & 41.0 & 42.4 & 79.7 & 84.2 & 54.9 & 23.4 & 78.8 & 54.4 & 46.0 & 75.9 & 89.6 & 22.8 & 51.3 & 72.2 & $\bm{66.1}$ & 74.9 & 76.0 & 63.0\\
  \methodname-OB-G-Ens.+FRCNN & 83.8 & $\bm{85.1}$ & \underline{$\bm{65.5}$} & 43.1 & \underline{$\bm{50.8}$} & $\bm{83.2}$ & \underline{$\bm{85.3}$} & 59.3 & 28.5 & \underline{$\bm{82.2}$} & \underline{$\bm{57.4}$} & 50.7 & \underline{$\bm{85.0}$} & \underline{$\bm{92.0}$} & $\bm{27.9}$ & $\bm{54.2}$ & 72.2 & 65.9 & \underline{$\bm{77.6}$} & \underline{$\bm{82.1}$} & \underline{$\bm{66.6}$}\\
\hline
\end{tabular}
}
\end{center}
\label{table:voc_2007_corloc}
\end{table*}
}

\subsubsection{The influence of the IoU threshold $I_{t}$}
\label{sec:influence_it}

Here we discuss the influence of the IoU threshold $I_{t}$ defined in Section~\ref{sec:fpcc}~(2) and Eq.~(\ref{equ:e}).
From Fig.~\ref{fig:ablation_threshold}, we see that setting $I_{t}$ to $0.4$ obtains the best performance.
Therefore, we set $I_{t}$ to $0.4$ for the other experiments.

{\addtolength{\tabcolsep}{-2pt}
\begin{table*}[t]
\caption{Results (AP in $\%$) for different methods on the VOC 2012 test set.
See Section~\ref{sec:compar} for the definitions of the \methodname-based methods.}
\begin{center}
\resizebox{\linewidth}{!}{
\begin{tabular}{|@{}L{3.15cm}|*{20}{x}|x|}
   \hline
   Method & aero & bike & bird & boat & bottle & bus & car & cat & chair & cow & table & dog & horse & mbike & person & plant & sheep & sofa & train & tv & mAP \\
   \hline
   \hline
   WSDDN+context \cite{Ref:Kantorov2016} & 64.0 & 54.9 & 36.4 & 8.1 & 12.6 & 53.1 & 40.5 & 28.4 & 6.6 & 35.3 & $\bm{34.4}$ & $\bm{49.1}$ & 42.6 & 62.4 & $\bm{19.8}$ & 15.2 & 27.0 & 33.1 & 33.0 & 50.0 & 35.3\\
   WCCN \cite{Ref:Diba2017} & - & - & - & - & - & - & - & - & - & - & - & - & - & - & - & - & - & - & - & - & 37.9\\
   Jie~\etal\ \cite{Ref:Jie2017} & 60.8 & 54.2 & 34.1 & 14.9 & 13.1 & 54.3 & 53.4 & $\bm{58.6}$ & 3.7 & 53.1 & 8.3 & 43.4 & 49.8 & 69.2 & 4.1 & 17.5 & 43.8 & 25.6 & $\bm{55.0}$ & 50.1 & 38.3\\
   \hline
   \methodname-OB-G-VGG$\_$M & 63.2 & 58.0 & 37.8 & 19.6 & 18.9 & 48.9 & 49.5 & 27.9 & 5.6 & 45.5 & 13.7 & 45.8 & 53.4 & 65.9 & 8.2 & 20.7 & 40.4 & 41.7 & 36.9 & 50.5 & 37.6\\
   \methodname-OB-G-VGG16 & 58.2 & 66.0 & 41.8 & 24.8 & 27.2 & $\bm{55.7}$ & 55.2 & 28.5 & $\bm{16.6}$ & 51.0 & 17.5 & 28.6 & 49.7 & 70.5 & 7.1 & 25.7 & $\bm{47.5}$ & 36.6 & 44.1 & $\bm{59.2}$ & 40.6\\
   \hline
   \methodname-OB-G-Ens. & 63.4 & 64.2 & 44.2 & 25.6 & 26.4 & 54.5 & 55.1 & 30.5 & 11.6 & 51.0 & 15.8 & 39.4 & 55.9 & 70.7 & 8.2 & $\bm{26.3}$ & 46.9 & 41.3 & 44.1 & 57.7 & 41.6\\
   \methodname-OB-G-Ens.+FRCNN & $\bm{69.0}$ & $\bm{71.3}$ & $\bm{56.1}$ & $\bm{30.3}$ & $\bm{27.3}$ & 55.2 & $\bm{57.6}$ & 30.1 & 8.6 & $\bm{56.6}$ & 18.4 & 43.9 & $\bm{64.6}$ & $\bm{71.8}$ & 7.5 & 23.0 & 46.0 & $\bm{44.1}$ & 42.6 & 58.8 & $\bm{44.2}$\\
\hline
\end{tabular}
}
\end{center}
\label{table:voc_2012_map}
\end{table*}
}

{\addtolength{\tabcolsep}{-2pt}
\begin{table*}[t]
\caption{Results (CorLoc in $\%$) for different methods on the VOC 2012 trainval set.
See Section~\ref{sec:compar} for the definitions of the \methodname-based methods.}
\begin{center}
\resizebox{\linewidth}{!}{
\begin{tabular}{|@{}L{3.15cm}|*{20}{x}|x|}
   \hline
   Method & aero & bike & bird & boat & bottle & bus & car & cat & chair & cow & table & dog & horse & mbike & person & plant & sheep & sofa & train & tv & mean \\
   \hline
   \hline
   WSDDN+context \cite{Ref:Kantorov2016} & 78.3 & 70.8 & 52.5 & 34.7 & 36.6 & 80.0 & 58.7 & 38.6 & 27.7 & 71.2 & 32.3 & 48.7 & 76.2 & 77.4 & 16.0 & 48.4 & 69.9 & 47.5 & 66.9 & 62.9 & 54.8\\
   Jie~\etal\ \cite{Ref:Jie2017} & 82.4 & 68.1 & 54.5 & 38.9 & 35.9 & $\bm{84.7}$ & 73.1 & $\bm{64.8}$ & 17.1 & 78.3 & 22.5 & 57.0 & 70.8 & 86.6 & 18.7 & 49.7 & 80.7 & 45.3 & 70.1 & 77.3 & 58.8\\
   \hline
   \methodname-OB-G-VGG$\_$M & 82.1 & 81.6 & 67.0 & 48.4 & 42.7 & 78.6 & 73.1 & 40.1 & 24.7 & 82.2 & 42.1 & $\bm{61.6}$ & 83.4 & 87.1 & 21.7 & 53.7 & 80.7 & $\bm{64.9}$ & 63.8 & 78.5 & 62.9\\
   \methodname-OB-G-VGG16 & 77.2 & 83.0 & 62.1 & 55.0 & 49.3 & 83.0 & 75.8 & 37.7 & $\bm{43.2}$ & 81.6 & $\bm{46.8}$ & 42.9 & 73.3 & 90.3 & 21.4 & 56.7 & 84.4 & 55.0 & 62.9 & 82.5 & 63.2\\
   \hline
   \methodname-OB-G-Ens. & 82.7 & 84.8 & 69.5 & 56.4 & 49.2 & 80.0 & 76.2 & 39.4 & 35.4 & 82.8 & 45.2 & 51.4 & 82.2 & 89.6 & 21.9 & 59.0 & 83.4 & 62.9 & 66.4 & 82.4 & 65.0\\
   \methodname-OB-G-Ens.+FRCNN & $\bm{86.7}$ & $\bm{86.7}$ & $\bm{74.8}$ & $\bm{56.8}$ & $\bm{53.8}$ & 84.2 & $\bm{80.1}$ & 42.0 & 36.4 & $\bm{86.7}$ & 46.5 & 54.1 & $\bm{87.0}$ & $\bm{92.7}$ & $\bm{24.6}$ & $\bm{62.0}$ & $\bm{86.2}$ & 63.2 & $\bm{70.9}$ & $\bm{84.2}$ & $\bm{68.0}$\\
\hline
\end{tabular}
}
\end{center}
\label{table:voc_2012_corloc}
\end{table*}
}

\subsubsection{The influence of the number of $k$-means clusters}
\label{sec:influence_kmeans}

In previous experiments we set the number of $k$-means clusters described in the last paragraph of Section~\ref{sec:fpcc}~(2) to be $3$.
Here we set it to other numbers to explore its influence.
The results from other numbers of $k$-means clusters are mAP $40.2\%$ and CorLoc $59.3\%$ for $2$ clusters,
and mAP $40.7\%$ and CorLoc $59.6\%$ for $4$ clusters,
which are a little worse than the results from $3$ cluster.
Therefore, we set the number of $k$-means clusters to $3$ for the other experiments.

\subsubsection{The influence of the IoU threshold $I^{\prime}_{t}$}
\label{sec:influence_it1}

We also analyse the influence of $I^{\prime}_{t}$ defined in Section~\ref{sec:pcg} and the $5$-th line of Algorithm~\ref{alg:pcg}.
As shown in Fig.~\ref{fig:ablation_threshold1}, $I^{\prime}_{t}=0.5$ outperforms other choices.
Therefore, we set $I^{\prime}_{t}$ to $0.5$ for the other experiments.

{
\subsubsection{The influence of multi-scale training and testing}
\label{sec:influence_scale}

Previously our experiments are conducted based on five image scales for training and testing.
Here we show the influence of this multi-scale setting.
We train and test our method using a single image scale $600$
as the default scale setting of FRCNN \cite{Ref:Girshick2015}.
The single-scale results are mAP $37.4\%$ and CorLoc $55.5\%$
which are much worse than our multi-scale results (mAP $40.8\%$ and CorLoc $59.6\%$).
Therefore, we use five image scales as many WSOD networks \cite{Ref:Diba2017,Ref:Bilen2016,Ref:Kantorov2016}.

\subsubsection{The number of proposal cluster centers}
\label{sec:nk}

As we stated in Section~\ref{sec:fpcc}~(2),
the number of proposal cluster centers (\ie, $N^{k}$)
changes for each image in each training iteration.
Here we give some typical values of $N^{k}$.
In the beginning of training,
the proposal scores are very noisy
and thus the selected top ranking proposals to form graphs are scattered in images,
which results in dozens of proposal cluster centers for each image.
After some (about 3K) training iterations,
the proposal scores are more reliable
and our method finds 1$\sim$3 proposal cluster centers for each positive object class.
To make the training more stable in the beginning,
for each positive object class we empirically select at most five proposal cluster centers which have higher scores,
and the number of selected proposal cluster centers does not influence the performance much.

}

\subsection{Comparison with other methods}
\label{sec:compar}

Here we compare our best performed strategy \methodname-OB-G,
\ie, using graph-based method and treating clusters as bags to train the network online,
with other methods.

We first report our results for each class on VOC 2007 and 2012 in Table~\ref{table:voc_2007_map}, Table~\ref{table:voc_2007_corloc}, Table~\ref{table:voc_2012_map}, and Table~\ref{table:voc_2012_corloc}.
It is obvious that our method outperforms other methods \cite{Ref:Bilen2016,Ref:Kantorov2016} using single model VGG$\_$M or VGG16 (\methodname-OB-G+VGG$\_$M and \methodname-OB-G+VGG16 in tables.)
Our single model results even better than others by combining multiple different models (\eg, ensemble of models) \cite{Ref:Bilen2016,Ref:Li2016,Ref:Jie2017,Ref:Diba2017}.
Specially, our method obtains much better results compared with other two methods also using the same basic MIL network \cite{Ref:Bilen2016,Ref:Kantorov2016}.
Importantly, \cite{Ref:Bilen2016} also equips the weighted sum pooling with objectness measure of EdgeBox \cite{Ref:Zitnick2014} and the spatial regulariser, and \cite{Ref:Kantorov2016} adds context information into the network, both of which are more complicated than our basic MIL network.
We believe that our performance can be improved by choosing better basic MIL networks, like the complete network in \cite{Ref:Bilen2016} and using context information \cite{Ref:Kantorov2016}.
As reimplementing their method completely is non-trivial, here we only choose the simplest architecture in \cite{Ref:Bilen2016}.
Even in this simplified case, our method achieves very promising results.

\begin{table}[t]
\caption{Results (mAP in $\%$) for different methods on the ImageNet dataset.
See Section~\ref{sec:compar} for the definitions of the \methodname-based methods.}
\begin{center}
\footnotesize
\begin{tabular}{|l|c|}
  \hline
  Method & Results \\
  \hline
  \hline
  Ren \etal\ \cite{Ref:Ren2016} & 9.6 \\
  Li \etal\ \cite{Ref:Li2016} & 10.8 \\
  WCCN \cite{Ref:Diba2017} & 16.3 \\
  \hline
  \methodname-OB-G-VGG$\_$M & 14.4 \\
  \methodname-OB-G-VGG16 & 18.4 \\
  \methodname-OB-G-Ens. & 18.8 \\
  \methodname-OB-G-Ens.+FRCNN & $\bm{19.6}$ \\
  \hline
\end{tabular}
\end{center}
\label{table:imagenet}
\end{table}

Our results can also be improved by combing multiple models.
As shown in the tables, there are little improvements from the ensemble of the VGG$\_$M and VGG16 models (\methodname-OB-G-Ens. in tables).
Here we do the ensemble by summing up the scores produced by the two models.
Also, as mentioned in Section~\ref{sec:exp_setup},
similar to \cite{Ref:Li2016,Ref:Diba2017,Ref:Jie2017},
we train a FRCNN detector using top-scoring proposals produced by \methodname-OB-G-Ens. as groundtruths (\methodname-OB-G-Ens.+FRCNN in tables).
As we can see, the performance is improved further.

We then show results of our method on the large scale ImageNet detection dataset in Table~\ref{table:imagenet}.
We observe similar phenomenon that our method outperforms other methods by a large margin.

\begin{table}[t]
\caption{Results (mAP@0.5 and mAP@[.5, .95] in $\%$) of different methods on the MS-COCO dataset.
See Section~\ref{sec:compar} for the definitions of the \methodname-based methods.}
{
\begin{center}
\footnotesize
\begin{tabular}{|l|c|c|}
  \hline
  Method & mAP@0.5 & mAP@[.5, .95] \\
  \hline
  \hline
  Ge \etal\ \cite{Ref:Ge2018} & 19.3 & 8.9 \\
  \hline
  \methodname-OB-G-VGG$\_$M & 16.6 & 7.3 \\
  \methodname-OB-G-VGG16 & 19.4 & 8.5 \\
  \methodname-OB-G-Ens. & 19.5 & 8.6 \\
  \methodname-OB-G-Ens.+FRCNN & $\bm{19.6}$ & $\bm{9.2}$ \\
  \hline
\end{tabular}
\end{center}}
\label{table:coco}
\end{table}

\begin{figure}[t]
\begin{center}
   \includegraphics[width=\linewidth]{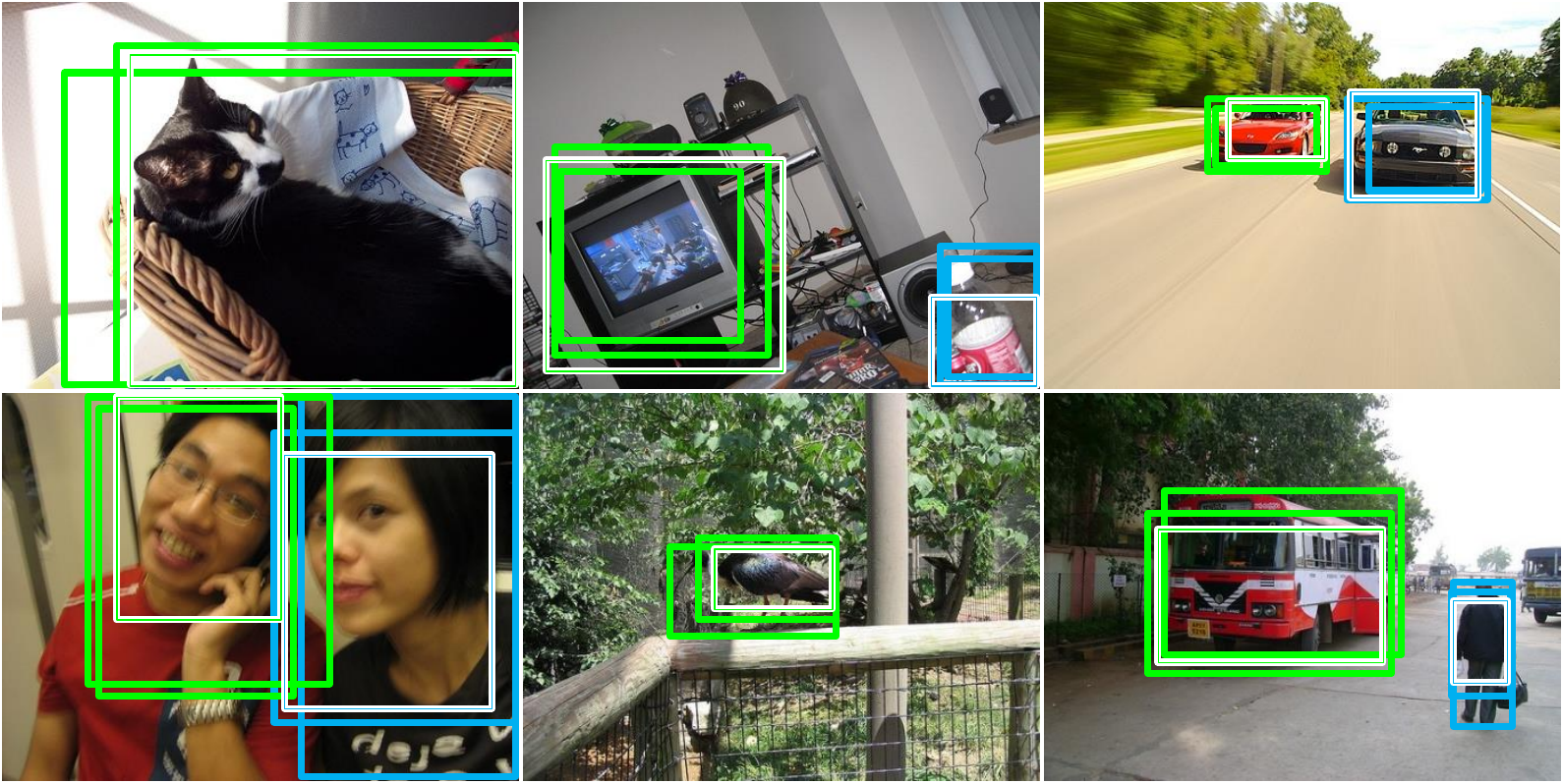}
\end{center}
   \caption{Some learned proposal clusters.
   The proposals with white are cluster centers.
   Proposals with the same color belong to the same cluster.
   We omit the background cluster for simplification.}
\label{fig:learned_clusters}
\end{figure}

\begin{figure*}[!t]
\begin{center}
   \includegraphics[width=\linewidth]{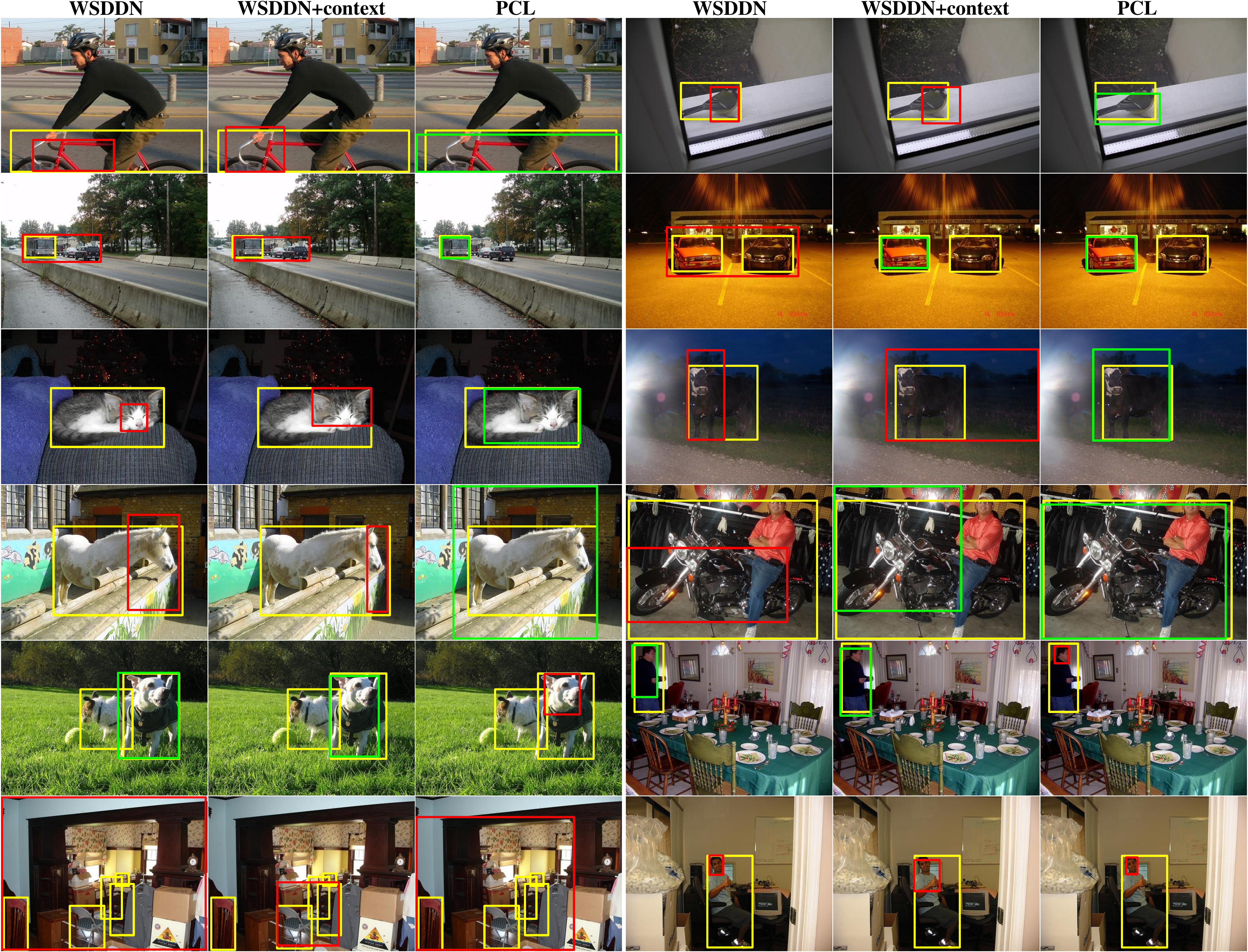}
\end{center}
   \caption{Some visualization comparisons among the WSDDN \cite{Ref:Bilen2016}, the WSDDN+context \cite{Ref:Kantorov2016}, and our method (\methodname)
   (in each image only the top-scoring box is shown).
   Green rectangle indicates success cases (IoU$>$0.5), red rectangle indicates failure cases (IoU$<$0.5), and yellow rectangle indicates groundtruths.
   The first four rows show examples that our method outperforms other two methods (with larger IoU).
   The fifth row shows examples that our method is worse than other two methods (with smaller IoU).
   The last row shows failure examples for both three methods.
   }
\label{fig:vis_qua}
\end{figure*}

\begin{figure*}[!t]
\begin{center}
   \includegraphics[width=\linewidth]{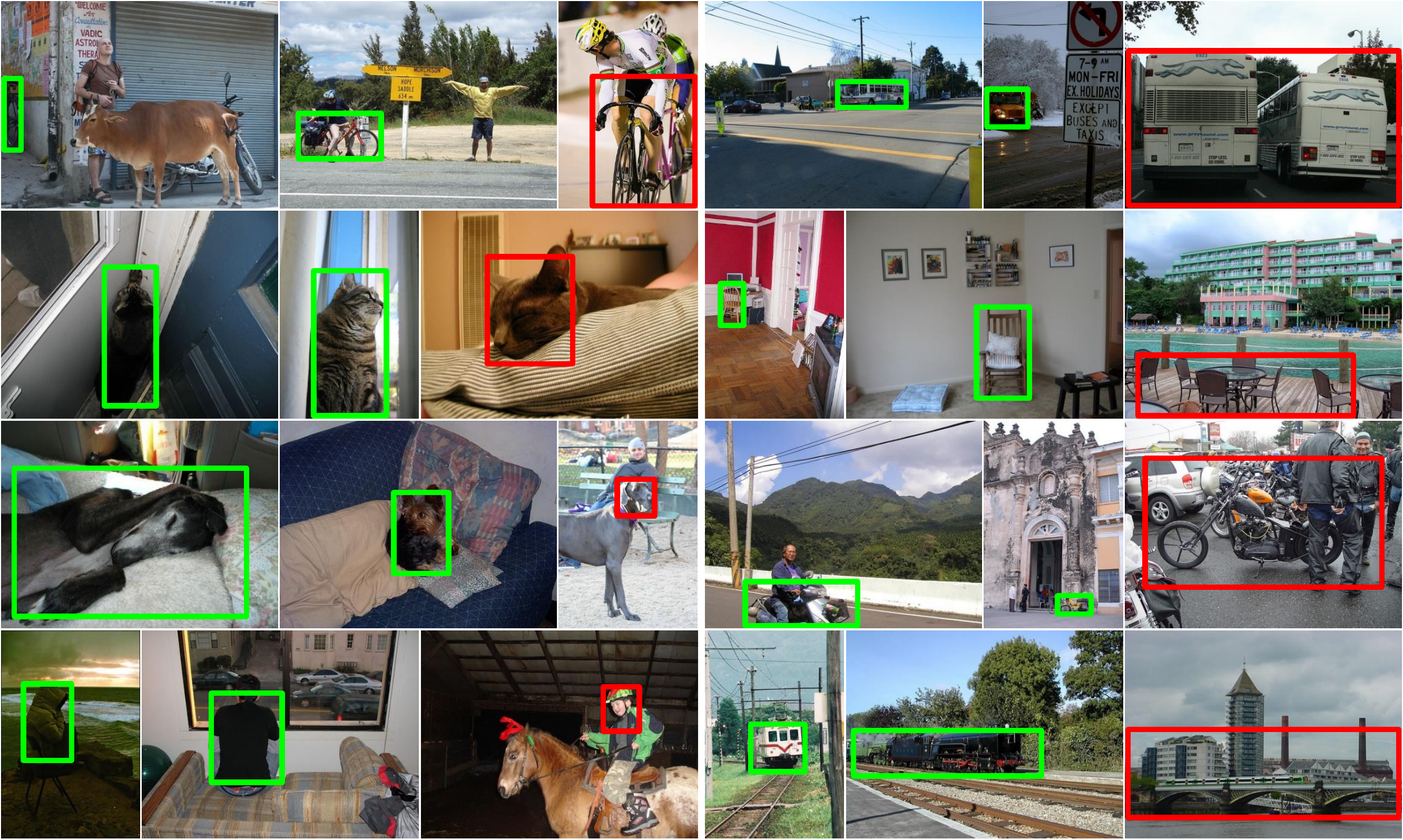}
\end{center}
   \caption{Some detection results for class bicycle, bus, cat, chair, dog, motorbike, person, and train
   (in each image only the top-scoring box is shown).
   Green rectangle indicates success cases (IoU$>$0.5), and red rectangle indicates failure cases (IoU$<$0.5).
   }
\label{fig:vis_final}
\end{figure*}

{
We finally report results of our method on MS-COCO in Table~\ref{table:coco}.
Our method obtains better performance than the recent work \cite{Ref:Ge2018}.
In particular, Ge \etal\ \cite{Ref:Ge2018} use the method proposed in our conference version paper \cite{Ref:Tang2017multiple} as a basic component.
We can expect to obtain better detection performance
through replacing our conference version method in \cite{Ref:Ge2018}
by our newly proposed method here,
which we would like to explore in the future.
}

\subsection{Qualitative results}
\label{sec:qua}

We first show some proposal clusters generated by our method in Fig.~\ref{fig:learned_clusters}.
As we can see, the cluster centers contain at least parts of objects
and are able to cover adaptive number of objects for each class.

We then show qualitative comparisons among the WSDDN \cite{Ref:Bilen2016}, the WSDDN+context \cite{Ref:Kantorov2016}, and our \methodname\ method, both of which use the same basic MIL network.
As shown in Fig.~\ref{fig:vis_qua}, we can observe that for classes such as bike, car, cat, \etc, our method tends to provide more accurate detections, whereas other two methods sometimes fails by producing boxes that are overlarge or only contain parts of objects (the first four rows in Fig.~\ref{fig:vis_qua}).
But for some classes such as person, our method sometimes fails by only detecting parts of objects such as the head of person (the fifth row in Fig.~\ref{fig:vis_qua}).
Exploiting context information sometimes help the detection (as in WSDDN+context \cite{Ref:Kantorov2016}), we believe our method can be further improved by incorporating context information into our framework.
All these three methods (actually almost all weakly supervised object detection methods) suffers from two problems:
producing boxes that not only contain the target object but also include their adjacent similar objects,
or only detecting parts of object for objects with deformation (the last row in Fig.~\ref{fig:vis_qua}).

We finally visualize some success and failure detection results on VOC 2007 trainval by \methodname-Ens.+FRCNN, as in Fig.~\ref{fig:vis_final}.
We observe similar phenomena as in Fig.~\ref{fig:vis_qua}.
Our method is robust to the size and aspect of objects, especially for rigid objects.
The main failures for these rigid objects are always due to overlarge boxes that not only contain objects, but also include adjacent similar objects.
For non-rigid objects like ``cat'', ``dog'', and ``person'', they often have great deformations,
but their parts (\eg, head of person) have much less deformation,
so our detector is still inclined to find these parts.
An ideal solution is yet wanted because there is still room for improvement.

{

\subsection{Runtime}
\label{sec:runtime}

\begin{table}[t]
\caption{Runtime comparisons between our method (``PCL'' in table) and our basic MIL network \cite{Ref:Bilen2016} (``Basic'' in table).}
{
\begin{center}
\footnotesize
\resizebox{\linewidth}{!}{
\begin{tabular}{| l | c c | c c |}
  \hline
  & \multicolumn{2}{c|}{PCL} & \multicolumn{2}{c|}{Basic} \\
  & VGG$\_$M & VGG16 & VGG$\_$M & VGG16\\
  \hline
  \hline
  Training (second/iteration) & 1.11 & 1.51  & 0.99 & 1.40 \\
  Testing (second/image) & 0.71 & 1.22 & 0.71 & 1.21 \\
  \hline
\end{tabular}
}
\end{center}
}
\label{table:time}
\end{table}

The runtime comparisons between our method and our basic MIL network \cite{Ref:Bilen2016} are shown in Table~\ref{table:time},
where the runtime of proposal generation is not considered.
As we can see,
although our method has more components than our basic MIL network \cite{Ref:Bilen2016},
our method takes almost the same testing time as it.
This is because all our output streams share the same proposal feature computations.
The small extra training computations of our method mainly
come from the procedures to find proposal cluster centers and generate proposal clusters.
Although with small extra training computations,
our method obtains much better detection results than the basic MIL network.

}

\section{Conclusion}
\label{sec:conclu}

In this paper, we propose to generate proposal clusters to learn refined instance classifiers for weakly supervised object detection.
We propose two strategies for proposal cluster generation and classifier refinement,
both of which can boost the performance significantly.
The classifier refinement is implemented by multiple output streams corresponding to some instance classifiers in multiple instance learning networks.
An online training algorithm is introduced to train the proposed network end-to-end for effectiveness and efficiency.
Experiments show substantial and consistent improvements by our method.
We observe that the most common failure cases of our algorithm are connected with the deformation of non-rigid objects.
In the future, we will concentrate on this problem.
In addition, we believe our learning algorithm has the potential to be applied in other weakly supervised visual learning tasks such as weakly supervised semantic segmentation.
We will also explore how to apply our method to these related applications.

\section*{Acknowledgements}
This work was supported by NSFC (No. 61733007, No. 61572207, No. 61876212, No. 61672336, No. 61573160), ONR with grant N00014-15-1-2356, Hubei Scientific and Technical Innovation Key Project, and the Program for HUST Academic Frontier Youth Team. The corresponding author of this paper is Xinggang Wang.

\ifCLASSOPTIONcaptionsoff
  \newpage
\fi



\bibliographystyle{IEEEtran}
\bibliography{egbib}
%



\begin{IEEEbiography}[{\includegraphics[width=1in,height=1.25in,clip]{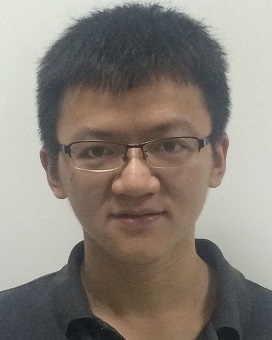}}]{Peng Tang}
received the B.S. degree in Electronics and Information Engineering from Huazhong University of Science and Technology (HUST) in 2014.
He is currently pursuing the Ph.D. degree in the School of Electronic Information and Communications at HUST,
and visiting the Department of Computer Science at Johns Hopkins University.
He was an intern at Microsoft Research Asia in 2017.
His research interests include image classification and object detection in images/videos.
\end{IEEEbiography}

\begin{IEEEbiography}[{\includegraphics[width=1in,height=1.25in,clip]{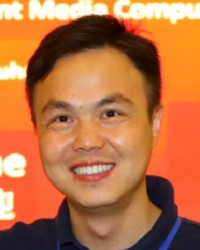}}]{Xinggang Wang}
is an assistant professor of School of Electronics Information and Communications of Huazhong University of Science and Technology (HUST). He received his Bachelor degree in communication and information system and Ph.D. degree in computer vision both from HUST. From May 2010 to July 2011, he was with the Department of Computer and Information Science, Temple University, Philadelphia, PA., as a visiting scholar. From February 2013 to September 2013, he was with the University of California, Los Angeles (UCLA), as a visiting graduate researcher. He is a reviewer of IEEE Trans on PAMI, IEEE Trans on Image Processing, IEEE Trans. on Cybernetics, Pattern Recognition, Computer Vision and Image Understanding, Neurocomputing, NIPS, ICML, CVPR, ICCV and ECCV etc. His research interests include computer vision and machine learning, especially object recognition.
\end{IEEEbiography}

\begin{IEEEbiography}[{\includegraphics[width=1in,height=1.25in,clip]{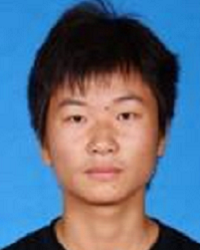}}]{Song Bai}
received the B.S. and Ph.D. degree in Electronics and Information Engineering from Huazhong University of Science and Technology (HUST), Wuhan, China in 2013 and 2018, respectively. He was with University of Texas at San Antonio (UTSA) and Johns Hopkins University (JHU) as a research scholar. His research interests include image retrieval and classification, 3D shape recognition, person re-identification, semantic segmentation and deep learning. More information can be found in his homepage: \url{http://songbai.site/}.
\end{IEEEbiography}

\begin{IEEEbiography}[{\includegraphics[width=1in,height=1.25in,clip]{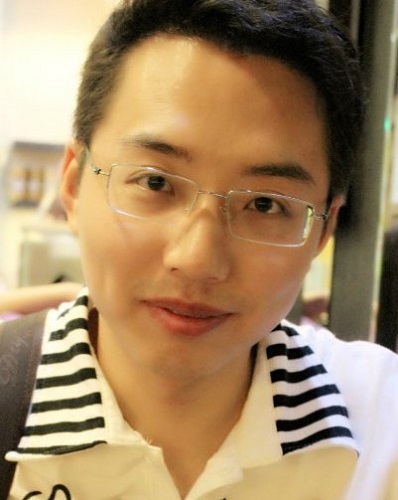}}]{Wei Shen}
received his B.S. and Ph.D. degree both in Electronics and Information Engineering from the Huazhong University of Science and Technology (HUST), Wuhan, China, in 2007 and in 2012.
From April 2011 to November 2011, he worked in Microsoft Research Asia as an intern. In 2012, he joined School of Communication and Information Engineering, Shanghai University as an Assistant Professor. From 2017, he became an Associate Professor.
He is currently visiting Department of Computer Science, Johns Hopkins University.
His current research interests include random forests, deep learning, object detection and segmentation.
\end{IEEEbiography}

\begin{IEEEbiography}[{\includegraphics[width=1in,height=1.25in,clip]{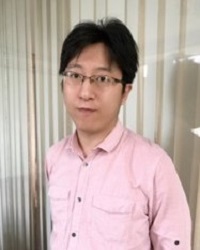}}]{Xiang Bai}
received his B.S., M.S., and Ph.D. degrees from the Huazhong University of Science and Technology (HUST), Wuhan, China, in 2003, 2005, and 2009, respectively, all in electronics and information engineering. He is currently a Professor with the School of Electronic Information and Communications, HUST. He is also the Vice-director of the National Center of Anti-Counterfeiting Technology, HUST. His research interests include object recognition, shape analysis, scene text recognition and intelligent systems. He serves as an associate editor for Pattern Recognition, Pattern Recognition Letters, Neurocomputing and Frontiers of Computer Science.
\end{IEEEbiography}

\begin{IEEEbiography}[{\includegraphics[width=1in,height=1.25in,clip]{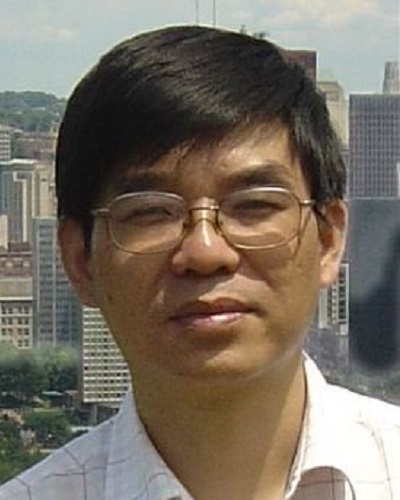}}]{Wenyu Liu}
received the B.S. degree in Computer Science from Tsinghua University, Beijing, China, in 1986, and the M.S. and Ph.D. degrees, both in Electronics and Information Engineering, from Huazhong University of Science and Technology (HUST), Wuhan, China, in 1991 and 2001, respectively. He is now a professor and associate dean of the School of Electronic Information and Communications, HUST. His current research areas include computer vision, multimedia, and machine learning. He is a senior member of IEEE.
\end{IEEEbiography}

\begin{IEEEbiography}[{\includegraphics[width=1in,height=1.25in,clip]{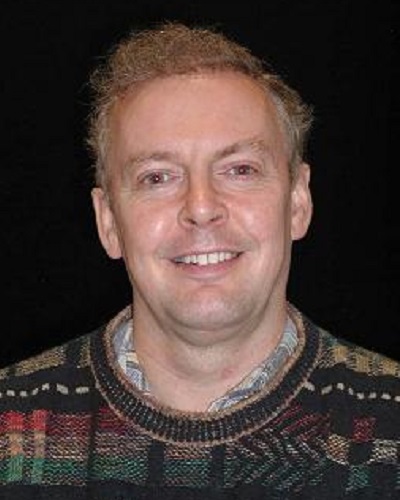}}]{Alan Yuille}
received the B.A. degree in mathematics from the University of Cambridge in 1976,
and the Ph.D. degree in theoretical physics from Cambridge in 1980.
He then held a post-doctoral position with the Physics Department, University of Texas, Austin,
and the Institute for Theoretical Physics, Santa Barbara.
He then became a Research Scientists with the Artificial Intelligence Laboratory, MIT, from 1982 to 1986,
and followed this with a faculty position in the division of applied sciences, Harvard,
from 1986 to 1995, rising to the position of an associate professor.
From 1995 to 2002, he was a Senior Scientist with the Smith-Kettlewell Eye Research Institute in San Francisco.
From 2002 to 2016, he was a Full Professor with the Department of Statistics, UCLA, with joint appointments in Psychology, Computer Science, and Psychiatry.
In 2016, he became a Bloomberg Distinguished Professor of cognitive science and computer science with Johns Hopkins University.
He received the Marr Prize and the Helmholtz Prize.
\end{IEEEbiography}

%








\end{document}